\documentclass{elsarticle} \usepackage[ruled]{algorithm}
\usepackage{algorithmic} \usepackage{amssymb} \usepackage{booktabs}
\usepackage{framed} \usepackage[margin=3cm,nohead]{geometry}
\usepackage{hyperref} \hypersetup{pdfauthor={Name}} \usepackage{mathtools}
\usepackage{caption,graphicx,newfloat} \usepackage{subfig} \usepackage{xcolor}
\usepackage[utf8]{inputenc}

\DeclareMathOperator*{\argmin}{arg\,min} \DeclareCaptionType{Algorithm}
\makeatletter \let\oldtheequation\theequation
\renewcommand\tagform@[1]{\maketag@@@{\ignorespaces#1\unskip\@@italiccorr}}
\renewcommand\theequation{(\oldtheequation)} \makeatother

\newcommand{\ah}[1]{\textcolor{black}{#1}}
\newcommand{\rv}[1]{\textcolor{black}{#1}}

\begin{document} 
\begin{frontmatter} 
    \title{A deep learning driven pseudospectral PCE based FFT homogenization 
    algorithm for complex microstructures} 
\author{Alexander Henkes\corref{cor1} \fnref{label1, label2}}
\ead{a.henkes@tu-braunschweig.de} \cortext[cor1]{Corresponding author}
\fntext[label2]{https://orcid.org/0000-0003-4615-9271} 
\author{Ismail Caylak \fnref{label1}} 
\author{Rolf Mahnken \fnref{label1}} 
\affiliation[label1]{
    organization={
    Chair of Engineering Mechanics, University of Paderborn},
    addressline={Warburger Str. 100}, 
    city={Paderborn}, 
    postcode={33098}, 
    country={Germany}}

    \begin{abstract} This work is directed to uncertainty quantification of
        homogenized effective properties for composite materials with complex,
        three dimensional microstructure. The uncertainties arise in the
        material parameters of the single constituents as well as in the fiber
        volume fraction. They are taken into account by multivariate random
        variables. Uncertainty quantification is achieved by an efficient
        surrogate model based on pseudospectral polynomial chaos expansion and
        artificial neural networks. An artificial neural network is trained on
        synthetic binary voxelized unit cells of composite materials with
        uncertain three dimensional microstructures, uncertain linear elastic
        material parameters and different loading directions. The prediction
        goals of the artificial neural network are the corresponding effective
        components of the elasticity tensor, where the labels for training are
        generated via a fast Fourier transform based numerical homogenization
        method. The trained artificial neural network is then used as a
        deterministic solver for a pseudospectral polynomial chaos expansion
        based surrogate model to achieve the corresponding statistics of the
        effective properties. Three numerical examples deal with the comparison
        of the presented method to the literature as well as the application to
    different microstructures. It is shown, that the proposed method is able to
predict central moments of interest while being magnitudes faster to evaluate
than traditional approaches.  \end{abstract}

    \begin{graphicalabstract}
    \includegraphics[width=\textwidth]{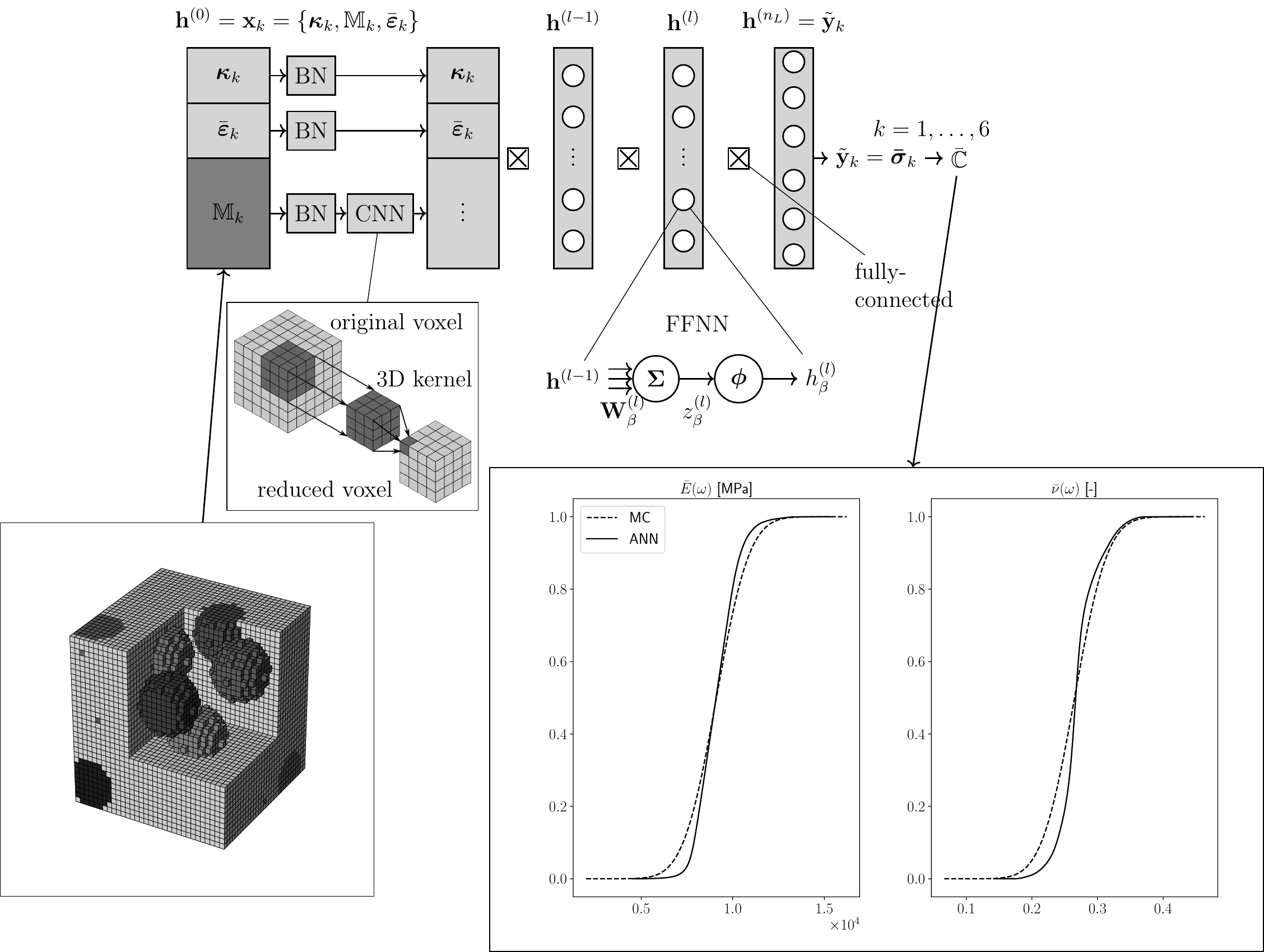}
\end{graphicalabstract}

    \begin{highlights} \item A deep learning algorithm for homogenization of
        uncertain complex microstructures is proposed \item A neural network is
        trained on three dimensional microstructures discretized by voxels and
        homogenized with FFT \item The geometry of the microstructures
        considered as well as their corresponding material parameters are
        modeled as random variables \item Uncertainty quantification is carried
        out by pseudospectral polynomial chaos expansion, utilizing the trained
    neural network as efficient solver \item Several numerical examples compare
the performance of the proposed method with respect to standard methods such as
FEM and Monte Carlo \end{highlights}


\end{frontmatter}
\section{Introduction} \label{sec:Introduction} Composite materials consist of
multiple constituents on the micro scale. This leads to a heterogeneous
microstructure in the sense of geometry as well as material behavior. In this
work, composites with two constituents of linear elastic material are discussed.
One constituent acts as a matrix, whereas the second is embedded in the former
as inclusions. The geometry of the inclusions can either be long fibers, short
fibers or particles \cite{aboudi2012micromechanics}. The microstructure and the
material behavior of its constituents determine the overall effective material
behavior of the composite on the macro scale. To obtain the effective macro
properties, homogenization techniques like analytical mean-field
\cite{bohm_short_nodate} or numerical full-field approaches are used, where the
later takes the microstructure into account explicitly. For complex
microstructures such as short fiber inclusions, analytical mean-field methods
perform poorly \cite{muller2015homogenization}. Full-field approaches on the
other hand can deal with arbitrary microstructures
\cite{muller2015homogenization}.

Microstructures as well as the material properties of the composite and
underlying constituents are subjected to uncertainties, grounded in either
intrinsic variety or induced by the manufacturing or measuring methods
\cite{kennedy2001bayesian, der2009aleatory}. This is especially true for short
fiber reinforced materials, where not only the microstructures and inclusion
geometries are very complex, but also the measurement and imaging is challenging
and dependent on many factors \cite{hiriyur2011uncertainty}. Uncertainties on
the micro scale lead to uncertain effective properties of the composite at the
macro scale, which requires uncertainty quantification (UQ).

\rv{This works follows the framework of \cite{sudret_surrogate_2017}, where UQ 
consists of three
steps.} First, the sources of uncertainties (e.g. microstructures and material
parameters) need to be quantified, e.g. by defining distributions of random
variables based on measurements and expert knowledge. Second, an appropriate
computational model of the problem studied needs to be defined. At last, the
uncertainties are propagated through the model to achieve the distribution of
the output, the so called quantity of interest (QoI). Universal techniques for
this purpose are Monte Carlo-type methods (MC) including simple Monte Carlo and
Latin Hyper-cube sampling \cite{hurtado_monte_1998}. Although they can be used
on existing computational models by simply sampling from the input distribution
and repeatedly running it, they suffer from low efficiency due to slow
convergence \cite{noauthor_monte_nodate}.

To address this problem, so called surrogate models can be used instead
\cite{sudret_surrogate_2017}. These apply some kind of dimension reduction e.g.
on the input parameters. Examples are polynomial chaos expansion (PCE, 
\cite{ghanem_stochastic_2003, xiu_wiener--askey_2002}), low-rank
tensor representations \cite{vondvrejc2020fft}, support-vector machines and
radial basis functions. 

\ah{
    Uncertainties in the context of continuum micromechanics were studied 
    extensively in e.g. \cite{soize2008tensor}. Here, a generic meso-scale
    probability model for a large class of random anisotropic elastic
microstructures in the context of tensor-valued random fields is proposed, which
is not limited to a specific microstructure defined by its constituents. The
stochastic boundary value problem is solved using the stochastic finite element
method (FEM). In \cite{cottereau2011localized}, a dualistic deterministic / stochastic
method is considered, utilizing two models coupled in an Arlequin framework. 
A framework using bounded random matrices for constitutive modeling is proposed
in \cite{noshadravan2013validation}, which results in a surrogate, which can be
calibrated to experimental data. In \cite{guilleminot2013stochastic}, a new
class of generalized nonparametric probabilistic models for matrix-valued
non-Gaussian random fields is investigated, where the focus is on random field
taking  values in some subset of the set of real symmetric positive-definite
matrices presenting sparsity and invariance with respect to given orthogonal
transformations. An approach using stochastic potential in combination with
a polynomial chaos representation for nonlinear constitutive equations of
materials in the context of random microstrucutral geometry and high dimensional
random parameters is given in \cite{clement2013uncertainty}.
}\rv{PCE was further used  in
homogenization by \cite{tootkaboni2010multi} and
\cite{caylak_polymorphic_nodate}. The later proposed an intrusive PCE in
combination with FEM based full-field homogenization
\cite{miehe_computational_2002} to determine uncertain effective material
properties of transversely linear elastic fiber reinforced composites.} The
intrusive approach to PCE uses Galerkin projection, where the FEM algorithm
needs to be reformulated using PC arithmetic \cite{debusschere2004numerical}.

While in \cite{caylak_polymorphic_nodate} cylindrical single fiber inclusions
are homogenized, in this work the uncertain homogenization method is adjusted to
more complex microstructures. For this it is convenient to replace the FEM
homogenization scheme by a fast Fourier transform (FFT) 
\rv{\cite{schneider2021review} based on the FFT Galerkin approach by
\cite{moulinec_fft-based_1995, de_geus_finite_2017}.} This method uses voxels 
and \rv{is more efficient in terms of CPU time and memory requirement than FEM,
as shown in \cite{cruzado2021variational} and
    \cite{VONDREJC2020112585}.} To 
adapt PCE to FFT, the Galerkin projection is replaced by the pseudospectral PCE
\cite{xiu_ecient_2007} to avoid PC arithmetics. Pseudospectral PCE uses
numerical integration techniques to obtain the PC coefficients. From this
surrogate model, the central moments of the QoI can be calculated. The
pseudospectral approach is called a non-intrusive method, as only repeated
deterministic solutions from the solver are needed instead of using PC
arithmetic. Here, the computational bottleneck is the deterministic solver
\cite{ghanem2017polynomial}.

To reduce the computational effort of the deterministic solution, data driven
machine learning models trained on the deterministic solver can be used to
replace the original model \cite{bock_review_2019}. Popular approaches are
decision trees, random forest ensembles, radial basis functions and support
vector machines. While relatively easy to train, they are limited to either
linear approximations or suffer from excessive parameters, especially in the
case of large three dimensional image data \cite{aggarwal_neural_2018}. An
alternative approach is the use of artificial neural networks (ANN). ANN,
especially in the context of deep learning, gained much attention, mainly
because of its impact in fields like computer vision, speech recognition and
autonomous driving \cite{bock_review_2019}. The universal approximation theorem
\cite{hornik_multilayer_1989} states, that ANN can learn any Borel measurable
function if the network has enough units. Additionally, there exist special
architectures to effectively handle large image data. There are already several
applications of ANN to computational continuum mechanics. For an overview the
reader should refer to \cite{bock_review_2019}. ANN were successfully used as
surrogate models in the context of MC based uncertainty quantification of
elliptic differential equations \cite{tripathy_deep_2018} and MC application on
three dimensional homogenization \cite{rao_three-dimensional_2020}. While there
were some applications to homogenization, these were limited to either two
dimensional microstructures \cite{yang_deep_2018}, uniaxial strain
\cite{beniwal_deep_2019} or fixed material parameters \cite{ye_deep_2019,
frankel_predicting_2019}.

To the authors knowledge there was no attempt to apply ANN to homogenization of
multiple three dimensional microstructures with uncertain material parameters
and different loading directions.  This work intends to close this gap by
establishing a complete homogenization technique based on ANN. Due to the fast
evaluation time of trained ANN, this homogenization technique is very efficient,
as shown in this work. The ANN can then be used to carry out otherwise expensive
stochastic investigations like pseudospectral PCE, where multiple deterministic
solutions are necessary. In summary the key objectives and contributions of this
work are: \begin{itemize} \item \textbf{Deep Learning homogenization algorithm:}
        An ANN is trained to homogenize uncertain three dimensional complex
        microstructures with uncertain material parameters subjected to
        different loading directions.  \item \textbf{FFT Label Generation:} FFT
        is used to provide the labels needed for training of the ANN. This is
        done on voxel based three dimensional microstructures without meshing
    \item \textbf{UQ with pseudospectral PCE:} A pseudospectral PCE using an ANN
        trained on FFT is used for UQ of the uncertain effective elasticity
        tensor of complex microstructures \end{itemize}

The rest of the paper is structured as follows. In \autoref{sec:Preliminaries}
the theoretical base of pseudospectral PCE, FFT based homogenization and ANN is
given. \autoref{sec:Algorithmic_Procedure} is concerned with the proposed
homogenization model of using ANN in UQ in the context of pseudospectral PCE.
Here the general problem formulation, data creation, network topology and
training of the ANN as well as the UQ using pseudospectral PCE are discussed.
\autoref{sec:Examples} consists of three numerical experiments. One aims to
compare the proposed method with \cite{caylak_polymorphic_nodate}, whereas the
second is dedicated to a more complex microstructure, showing also the
computational efficiency of the ANN accelerated approach to UQ. The third
example investigates the abillity of the ANN to generalize to unseen
microstructures. The paper closes with a summary and an outlook in
\autoref{sec:Summary_Outlook}.

\section{Preliminaries} \label{sec:Preliminaries} This section gives essential
prerequisites of random variables, uncertainty quantification, deep learning and
numerical homogenization, which are used in the proposed algorithm in
\autoref{sec:Algorithmic_Procedure}.  \subsection{Random variables and
uncertainty quantification} \label{sec:random_uq} Let $(\Omega, \Sigma, P)$ be a
probability space \cite{xiu2010numerical} with sample space $\Omega$,
$\sigma$-algebra $\Sigma$, and a probability measure $P$ on $\Omega$. A
multivariate random input variable $\mathbf{X}$ is defined by the map
\begin{equation} \mathbf{X}: \left\{\begin{aligned} &\Omega&\to
            &\quad\mathcal{D}_{\mathbf{X}} \subset \mathbb{R}^{n_x} \\
            &\omega&\mapsto &\quad\mathbf{X}(\omega) = \mathbf{x},
\label{eq:input} \end{aligned}\right.  \end{equation} where
$\mathcal{D}_{\mathbf{X}} \subset \mathbb{R}^{n_{x}}$ is an $n_x$ dimensional
vector-space of realizations $\mathbf{x}$ of elementary events $\omega \in
\Omega$ by the map $\mathbf{X}$. A model $\mathcal{M}$ then takes multivariate
random input variables $\mathbf{X}(\omega)$ and maps them to multivariate random
variables $\mathbf{Y}(\omega)$ defined by the relation \begin{equation}
    \mathcal{M}: \left\{\begin{aligned}
            &\mathcal{D}_{\mathbf{X}}&\to&\quad\mathcal{D}_{\mathbf{Y}} \subset
            \mathbb{R}^{n_{y}} \\
            &\mathbf{X}(\omega)&\mapsto&\quad\mathcal{M}(\mathbf{X}(\omega)) =
            \mathbf{Y}(\omega) = \mathbf{y}, \label{eq:multivariate_random}
\end{aligned}\right.  \end{equation} where $\mathcal{D}_{\mathbf{Y}} \subset
\mathbb{R}^{n_{y}}$ is an $n_y$ dimensional vector-space of realizations
$\mathbf{y}$ of elementary events $\omega \in \Omega$ by the composition
function $\mathbf{Y} = \mathcal{M} \circ \mathbf{X}: \Omega \to
\mathcal{D}_{\mathbf{Y}}$. The goal of uncertainty quantification (UQ) is to
characterize the distribution of the output vector $\mathbf{Y}(\omega)$ for a
given input vector $\mathbf{X}(\omega)$ in \autoref{eq:input}, thus propagating
input uncertainties through the model $\mathcal{M}$. Often the original model
$\mathcal{M}$ is computationally demanding. Therefore, surrogate models
$\tilde{\mathcal{M}}$ can be designed, which approximate the original model but
take less computing time to evaluate, as \begin{equation}    \tilde{\mathcal{M}}
    : \left\{\begin{aligned}
            &\mathcal{D}_{\mathbf{X}}&\to&\quad\tilde{\mathcal{D}}_{\mathbf{Y}}
            \subseteq {\mathcal{D}}_{\mathbf{Y}} \subset \mathbb{R}^{n_{y}}\\
            &\mathbf{X}(\omega)&\mapsto&\quad\tilde{\mathcal{M}}(\mathbf{X}(\omega))
            = \tilde{\mathbf{Y}}(\omega) = \tilde{\mathbf{y}},
\label{eq:surrogate1} \end{aligned}\right.  \end{equation} where
$\tilde{\mathcal{D}}_{\mathbf{Y}} \subseteq {\mathcal{D}}_{\mathbf{Y}} \subset
\mathbb{R}^{n_{y}}$ is an $n_y$ dimensional vector-space of realizations
$\tilde{\mathbf{y}}$ of $\omega \in \Omega$ by the composition function
$\tilde{\mathbf{Y}} = \tilde{\mathcal{M}} \circ \mathbf{X}: \Omega \to
\tilde{\mathcal{D}}_{\mathbf{Y}}$. Due to approximations, the datasets
$\tilde{\mathcal{D}}_{\mathbf{Y}}$ and ${\mathcal{D}}_{\mathbf{Y}}$ usually do
not coincide. This means, that two mappings of the same realization
$\mathbf{X}(\omega) = \mathbf{x}$ by the original model $\mathcal{M}$ in
\autoref{eq:multivariate_random} and a surrogate model $\tilde{\mathcal{M}}$ are
not the same. The distance between $\mathcal{M}$ and $\tilde{\mathcal{M}}$
defines an error function \cite{grimmett_probability_2020}
\begin{equation} \mathcal{E} = ||\mathcal{M}(\mathbf{X}(\omega)) -
    \tilde{\mathcal{M}}(\mathbf{X}(\omega))||_2 =
    \sqrt{\mathbb{E}[\mathcal{M}(\mathbf{X}(\omega)) -
    \tilde{\mathcal{M}}(\mathbf{X}(\omega))]^2}, \label{eq:error} \end{equation}
    with expectation $\mathbb{E}$ defined component wise as the integral 
    \rv{\begin{equation}
            {\displaystyle \mathbb{E}[\mathbf{X}(\omega)] =
                \operatorname {E} [(X_{1}(\omega), \ldots , X_{n_x}(\omega))]=
        (\mathbb{E} [X_{1}(\omega)], \ldots ,\mathbb{E} [X_{n_x}(\omega)])}, 
        \qquad
        \displaystyle \mathbb{E}[X(\omega)]=\int _{\Omega } X(\omega)
        \,\mathrm{d}{P},
    \label{eq:expectation} 
\end{equation}}over a
probability space $(\Omega, \Sigma, P)$. For a surrogate model
        $\tilde{\mathcal{M}}$ to have a small error $\mathcal{E}$ with respect
        to the original model $\mathcal{M}$ in \autoref{eq:multivariate_random},
        it must be calibrated on a dataset $\mathbb{D}$, which consists of
        evaluations of an original model $\mathcal{M}$ with regard to a finite
        number $n_s$ of realizations $\mathbf{x}_k \subset
        \mathcal{D}_{\mathbf{X}}$, called samples, of $\mathbf{X}(\omega)$ in
        \autoref{eq:input}, such that \begin{equation} \mathbb{D} =
        \left\{(\mathbf{x}_k, \mathbf{y}_k),\; k = 1, \dots, n_s\right\}.
    \label{eq:dataset} \end{equation}\rv{
    \subsection{Surrogate modelling by PCE}}
An example of a surrogate model as described in \autoref{eq:surrogate1} is
provided by a Polynomial Chaos Expansion $\tilde{\mathcal{M}}_{PCE}$ defined as
\begin{align} &1. \quad \tilde{\mathcal{M}}_{PCE}(\mathbf{X}(\omega)) =
    \tilde{\mathbf{Y}}(\omega) = \displaystyle\sum_{|\mathbf{i}| \leq n_{PCE}}
    \hat{\mathbf{Y}}_{\mathbf{i}}
    \Psi_{\mathbf{i}}(\mathbf{\theta}(\omega)),\qquad \text{where} \nonumber \\
    &2. \quad \hat{\mathbf{Y}}_{\mathbf{i}} =
    \displaystyle\frac{1}{\gamma_{\mathbf{i}}}
    \mathbb{E}\left[\tilde{\mathbf{Y}}(\omega)
    \Psi_{\mathbf{i}}(\mathbf{\theta}(\omega)) \right] =
    \displaystyle\frac{1}{\gamma_{\mathbf{i}}} \left\langle
    \tilde{\mathbf{Y}}(\omega), \Psi_{\mathbf{i}}(\mathbf{\theta}(\omega))
    \right\rangle = \displaystyle\frac{1}{\gamma_{\mathbf{i}}} \displaystyle\int
    \tilde{\mathbf{Y}}(\omega) \Psi_{\mathbf{i}}(\mathbf{\theta}(\omega))
    \mathrm{d}F_{\tilde{\mathbf{Y}}} \nonumber\\ &3. \quad \gamma_{\mathbf{i}} =
    \mathbb{E}\left[ \Psi_{\mathbf{i}}^2 \right] = \left\langle
    \Psi_{\mathbf{i}}, \Psi_{\mathbf{i}} \right\rangle \nonumber\\ &4. \quad
    F_{\tilde{\mathbf{Y}}}(\tilde{\mathbf{y}}) =
    P\left(\tilde{\mathbf{Y}}_{1}(\omega) \leq \tilde{\mathbf{y}}_{1}, \ldots,
    \tilde{\mathbf{Y}}_{n_y}(\omega) \leq \tilde{\mathbf{y}}_{n_y}\right)
\label{eq:surrogate_pce} \end{align} Here,
$\Psi_{\mathbf{i}}(\mathbf{\theta}(\omega))$ are orthonormal polynomials, which
act as basis functions of the expansion with truncation order $n_{PCE}$,
$\gamma_{\mathbf{i}}$ are normalization factors with inner product $\left\langle
\bullet, \bullet \right\rangle$, $\mathbf{\theta}(\omega)$ denote \ah{standard
normal}
distributed random variables and $F_{\tilde{\mathbf{Y}}}$ is the cumulative
distribution function (CDF) with respect to the multivariate random variable
$\tilde{\mathbf{Y}}(\omega)$. In \autoref{eq:surrogate_pce} and throughout the
remaining sections of this paper, $\mathbf{i}$ is a multi-index over the random
input space $\mathcal{D}_{\mathbf{X}} \subset \mathbb{R}^{n_{x}}$ in
\autoref{eq:input} \begin{equation} \mathbf{i} = (i_1,\dots,i_{n_{x}}) \in
\mathbb{N}_0^{n_{x}},\quad |\mathbf{i}| = i_1 + {\dots} + i_{n_x}.
\label{eq:mulind} \end{equation} The calibration of $\tilde{\mathcal{M}}_{PCE}$
from \autoref{eq:surrogate_pce} to the original model $\mathcal{M}$ in
\autoref{eq:multivariate_random} is performed by calculating the PC coefficients
$\hat{\mathbf{Y}}_{\mathbf{i}}$ 
\ah{in Eq. (7.2).}
To this end, the so called
pseudospectral approach to PCE uses a cubature rule for selection of the samples
in the dataset in \autoref{eq:dataset}\ah{, where the number of
    samples $n_s$ equals the number of cubature \rv{nodes} \rv{$\Theta$}, i.e. 
$n_s = n_q$, such that}
\ah{\begin{equation} \mathbb{D}_{cub} =
    \left\{(\mathbf{x}_k = \Theta_k, \dots, \mathbf{x}_{n_q} = \Theta_{n_q}),
    (\mathcal{M}(\Theta_k), \dots, \mathcal{M}(\Theta_{n_q})), \;
k=1,\dots,n_{s} = n_q\right\}. \label{eq:dataset_cub} \end{equation}}\rv{
    Here $\Theta_k,\; k=1, \dots, \ah{n_{q},}$ denotes the single 
    index form of the cubature nodes.} 
\ah{Accordingly with $\tilde{\mathbf{Y}}(\omega)$ the PC coefficients in
Eq. (7.2) are approximated as}
\begin{equation}
    \hat{\mathbf{Y}}_{\mathbf{i}} \approx
    \displaystyle\frac{1}{\gamma_{\mathbf{i}}}
    \displaystyle\sum_{j=1}^{n_{w}}{\mathcal{M}}(\Theta^{(j)})\Psi_{\mathbf{i}}
    (\Theta^{(j)})w(\Theta^{(j)}),
\label{eq:pseudo} \end{equation}\rv{where $\Theta^{(j)}, j=1, \dots, n_{w}$,
    represents the multi-index form of
    a cubature rule with dimension $n_x$ and weights $w(\Theta^{(j)})$, where
$n_{w}$ denotes the number of nodes and weights per dimension $n_{x}$}. 
The total number of \ah{cubature points 
$\mathbf{x}_{n_q}$ needed in
\autoref{eq:dataset_cub} is \begin{equation} n_q = (n_w)^{n_x},
\label{eq:no_solutions} \end{equation}}as the cubature rule with $n_w$ nodes
$\Theta^{(j)}$ and corresponding weights $w(\Theta^{(j)})$ must be carried out
over all dimensions $n_x$ of the input random vector space
$\mathcal{D}_{\mathbf{X}} \subset \mathbb{R}^{n_{x}}$ in \autoref{eq:input}.
\ah{The relation in \autoref{eq:no_solutions} holds true for equal number of
nodes and weights $n_w$ per dimension $n_x$, as it is the case in the reminder
of this work. Generally, $n_w$ can be chosen differently for each dimension.}
        
In this paper, the central moments of interest of the input $\mathbf{X}(\omega)$
and output $\mathbf{Y}(\omega)$ are the mean  \begin{equation}
    \boldsymbol{\mu}_{\mathbf{X}} =
    \mathbb{E}\left[\mathbf{X}(\omega)\right],\quad
\boldsymbol{\mu}_{\mathbf{X}} \in \rv{\mathbb{R}^{n_x}} 
\label{eq:mean} \end{equation}
    with expectation defined in \autoref{eq:expectation} and variance with
    standard deviation $\boldsymbol{\sigma}_{\mathbf{X}}$ \begin{equation}
        \boldsymbol{\sigma}_{\mathbf{X}}^2 = \text{var}(\mathbf{X}(\omega)) =
        \mathbb{E}\left[(\mathbf{X}(\omega) - \mathbb{E}\left[
            (\mathbf{X}(\omega) \right])^2 \right], \quad \boldsymbol{\sigma}
        \in \rv{\mathbb{R}^{n_x}_+}. 
    \label{eq:deviation} 
\end{equation} \rv{The
            mean $\boldsymbol{\mu}_{\mathbf{Y}}$ and the standard deviation
            $\boldsymbol{\sigma}_{\mathbf{Y}}$ can be calculated from the PC
            coefficients $\hat{\mathbf{Y}}_{\mathbf{i}}$ in \autoref{eq:pseudo}
        by the following relations} 
        \begin{equation}
                \boldsymbol{\mu}_{\mathbf{Y}} =
                \mathbb{E}\left[\mathcal{M}(\mathbf{X}(\omega))\right] \approx
                \mathbb{E}\left[\tilde{\mathcal{M}}(\mathbf{X}(\omega))\right]
                \approx \hat{\mathbf{Y}}_{\mathbf{0}} \in \rv{\mathbb{R}^{n_y}}
                \label{eq:moment1}
            \end{equation} 
            \begin{equation} \boldsymbol{\sigma}_{\mathbf{Y}}^2 =
                \text{var}({\hat{\mathbf{Y}}}_{\mathbf{i}}) =
                \mathbb{E}\left[({\mathcal{M}(\mathbf{X}(\omega)}) -
                \boldsymbol{\mu})^2\right] \approx \displaystyle\sum_{0 <
                |\mathbf{i}| \leq
            n_{PCE}}\left[\gamma_{\mathbf{i}}\hat{\mathbf{Y}}_{\mathbf{i}}^2
            \right] \in \rv{\mathbb{R}^{n_y}_+}.
        \label{eq:moment2} 
    \end{equation} For a more detailed treatment, the
        reader is referred to \cite{xiu2010numerical, grimmett_probability_2020,
        rozanov2013probability}.
 
 \rv{\subsection{Surrogate modeling by ANN}}
\label{sec:Deep_Learning}
\begin{figure}[htb]
\centering
\includegraphics[width=0.8\textwidth]{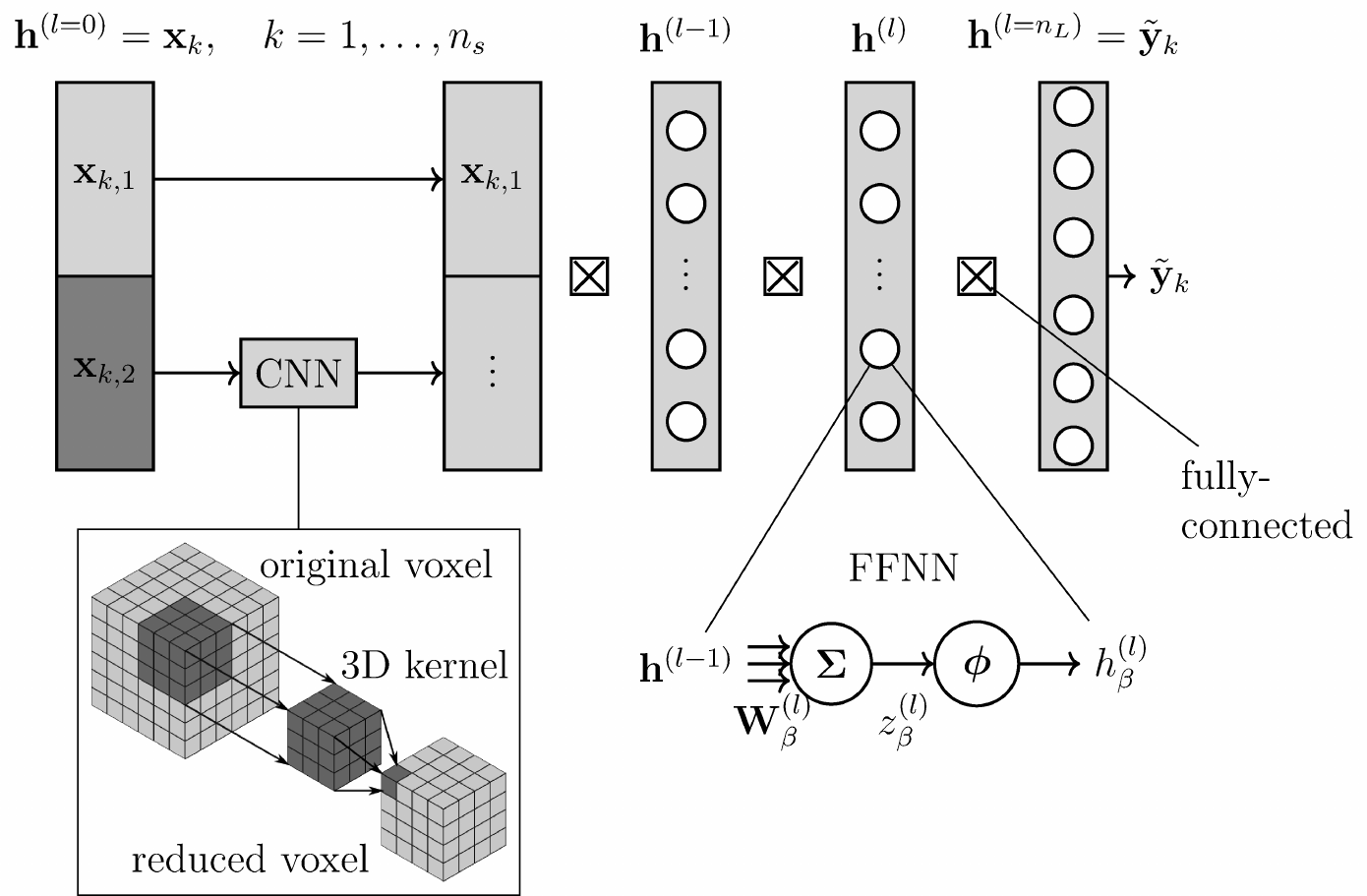}
        \caption{Schematic structure of an ANN in \autoref{eq:nn} with $n_L$ layers $\mathbf{h}^{(l)}$ and $n_u$ neural units $h_{\beta}^{(l)}$. An ANN with input ${\mathbf{x}}_k = \{{\mathbf{x}}_{k,1}, {\mathbf{x}}_{k,2}\}$, where ${\mathbf{x}}_{k,1}$ is a numerical vector and ${\mathbf{x}}_{k,2}$ consist of 3D voxel data. ${\mathbf{x}}_{k,2}$ is first processed by a CNN in \autoref{eq:CNN} and after concatenation with ${\mathbf{x}}_{k,1}$ feed into an FFNN in \autoref{eq:FFNN}. The resulting output is $\mathbf{h}^{(n_L)} =  \tilde{{\mathbf{y}}}_k$. The index $k = 1, \dots, n_s$, denotes the single samples, where $n_s$ is the number of samples.}
        \label{fig:ANN}
    \end{figure}

Deep learning using ANN is a subbranch of machine learning, which is a subbranch
of artificial intelligence \cite{goodfellow_deep_2016, aggarwal_neural_2018}. An
ANN, as visualized in \autoref{fig:ANN}, is a surrogate model
$\tilde{\mathcal{M}}_{ANN}$ in \autoref{eq:surrogate1}, which is built of $n_L$
layers $\mathbf{h}^{(l)}$ consisting of $n_u$ neural units $h_\beta^{(l)}$
            \begin{equation}
                \mathbf{h}^{(0)} = \mathbf{x}_{k},\quad\mathbf{h}^{(l)} = \left\{h_{\beta}^{(l)}, \beta = 1, \dots, n_{u}\right\},\quad\mathbf{h}^{(L)} = \tilde{{\mathbf{y}}}_{k},\quad l = 0, \dots, n_L.
                \label{eq:nn}
            \end{equation}
            Here, $\mathbf{h}^{(0)}$ is the first layer consisting of input
            samples ${\mathbf{x}}_{k}$ from a dataset $\mathbb{D}$ according to
            \autoref{eq:dataset}, where $k = 1, \dots, n_s$, denotes single
            samples of $\mathbb{D}$ and $n_s$ the number of samples.
            Furthermore, $\mathbf{h}^{(n_L)}$ is the last layer consisting of
            predictions $\tilde{\mathbf{y}}_{k} =
        \tilde{\mathcal{M}}_{ANN}(\tilde{\mathbf{x}}_{k})$ with respect to the input ${\mathbf{x}}_{k}$. For the neural units $h_{\beta}^{(l)}$ in \autoref{eq:nn}, several architectures exist, where the following are used throughout this paper:
\begin{itemize}
    \item Densely connected Feed Forward Neural Network (FFNN) 
    \begin{equation}
        h_{\beta}^{(l)} = \phi\left(\mathbf{W}_{\beta}^{(l)}\mathbf{h}^{(l-1)}\right) = \phi({z}_{\beta}^{(l)}),
        \label{eq:FFNN}
    \end{equation}
where $\mathbf{W}_{\beta}^{(l)}$ are weights for every unit $h_{\beta}^{(l)}$,
which are multiplied with the output of the preceding layer
$\mathbf{h}^{(l-1)}$. These weights are stored in a matrix $\mathbf{W} =
\left\{\mathbf{W}_{\beta}^{(l)}\right\}$ for the whole network. After
multiplication of weights $\mathbf{W}_{\beta}^{(l)}$ and output
$\mathbf{h}^{(l-1)}$, a nonlinear activation function $\phi$ is applied to the
product \rv{in \autoref{eq:FFNN}}. This is important for the network to be able to represent non-linearities. In this paper, \textit{rectified linear units} (ReLU) \cite{nair2010rectified}) are used
    \begin{equation}
        \phi(z_{\beta}^{(l)}) = \text{max}\left\{0, z_{\beta}^{(l)}\right\}.
        \label{eq:activation}
    \end{equation}
    \item Convolutional Neural Network (CNN) for three dimensional image data utilizing voxels, using a 3D kernel and the convolution operation 
    \begin{equation}
        h_{\beta}^{(l)} = \{h_{f,xyz}^{(l)}\} = \left\{\phi \left(\displaystyle\sum_{m=1}^{n_F^{(l-1)}}  \displaystyle\sum_{p=0}^{n_K^{(l)}-1}        \displaystyle\sum_{q=0}^{n_K^{(l)}-1} \displaystyle\sum_{r=0}^{n_K^{(l)}-1}  k_{fm,pqr}^{(l)} h_{m,(x+p)(y+q)(z+     r)}^{(l-1)} \right)\right\},
        \label{eq:CNN}
    \end{equation}
where $\{h_{f,xyz}^{(l)}\}$ denotes the output voxel of filter $f$, with $n_F$ being the number of filters, which are equivalent to the number of neural units $n_u$ of FFNN. $k_{fm,pqr}^{(l)}$ is the three dimensional kernel, which, similar to \autoref{eq:FFNN}, can be gathered in a matrix $\mathbf{K} = \{k_{fm,pqr}^{(l)}\}$ for the whole network and $n_K$ is the kernel dimension. 
\end{itemize}
The so called topology of an ANN is then the composition of neural units $h_{\beta}^{(l)}$ in \autoref{eq:nn} with their corresponding architectures FFNN in \autoref{eq:FFNN} and CNN in \autoref{eq:CNN} to a network \cite{sammut2011encyclopedia}. The calibration of the surrogate in \autoref{eq:nn}, also called training, consists of comparing the output data ${\mathbf{y}}_{k}$ of a model $\mathcal{M}$, defined in \autoref{eq:multivariate_random}, from a dataset $\mathbb{D}$ in \autoref{eq:dataset} consisting of predictions $\tilde{{\mathbf{y}}}_{k}(\mathbf{W}, \mathbf{K})$ from an ANN $\tilde{\mathcal{M}}_{ANN}$ in \autoref{eq:nn} with weights $\mathbf{W}$ in \autoref{eq:FFNN} or kernels $\mathbf{K}$ in \autoref{eq:CNN}. Consequently, the error function $\mathcal{E}$ in \autoref{eq:error} takes the form
\rv{
            \begin{equation}
                \mathcal{E} = \displaystyle\sum_k^{n_s}||\mathbf{y}_k - \tilde{\mathbf{y}}_k(\mathbf{W}, \mathbf{K}) ||_2 
                = \displaystyle\sum_k^{n_s}\sqrt{\mathbb{E}[\mathbf{y}_k -
                \tilde{\mathbf{y}}_k(\mathbf{W}, \mathbf{K})]^2},
                \label{eq:error2}
        \end{equation}where $k$ and $n_s$ are defined means of
    \autoref{eq:dataset}.} 
            This leads to a minimization problem, where an L2-regularization term $\mathcal{R}(\mathbf{W}, \mathbf{K})$ is added 
        \begin{equation}
            \argmin_{\mathbf{W}, \mathbf{K}}\; \left\{ \mathcal{E} + \mathcal{R}(\mathbf{W}, \mathbf{K}) \right\} = \argmin_{\mathbf{W}, \mathbf{K}}\; \left\{  ||\mathbf{y}_k - \tilde{\mathbf{y}}_k (\mathbf{W}, \mathbf{K})||_2 + \underbrace{\lambda_{L2} || \mathbf{W}, \mathbf{K} ||_2^2}_\text{$\mathcal{R}(\mathbf{W}, \mathbf{K})$} \right\},
            \label{eq:optimization}
        \end{equation}
        with regularization factor $\lambda_{L2}$. The regularization term
        $\mathcal{R}(\mathbf{W}, \mathbf{K})$ in \autoref{eq:optimization} is
        added to prevent overfitting, where the exact value for the
        regularization factor $\lambda_{L2}$ needs to be thoroughly tuned.
        Typically, $\lambda_{L2} \approx 0.01$ \rv{is used} \cite{tensorflow2015-whitepaper}. The weights $\mathbf{W}$ or kernels $\mathbf{K}$ are then updated iteratively by gradient descent of the error function $\mathcal{E}$ in \autoref{eq:error2} with respect to the weights or kernels, respectively
            \begin{equation}
                \mathbf{W} \leftarrow \mathbf{W} - \alpha \frac{\partial \mathcal{E}}{\partial \mathbf{W}},\quad \mathbf{K} \leftarrow \mathbf{K} - \alpha \frac{\partial \mathcal{E}}{\partial \mathbf{K}},
                \label{eq:update}
        \end{equation}
        where $\alpha$ denotes the gradient descent step width, also called learning rate. For further details the reader is referred to standard works such as \cite{zhang2019dive, goodfellow_deep_2016, aggarwal_neural_2018}.

\subsection{Numerical homogenization} \label{sec:FFT} The following section
gives an overview of numerical homogenization \rv{in micromechanics}. A
comprehensive treatment of the
topic can be found in \cite{bohm_short_nodate, aboudi2012micromechanics,
li_introduction_2008}.

\ah{Following the notation of 
\cite{vondrejc_fft-based_2014, de_geus_finite_2017, zeman2017finite, 
schneider2021review},} a microstructure is represented as a 
unit cell \begin{equation}
    \mathbb{M}(\zeta_i, \omega) = (0, \xi_1) \otimes (0, \xi_2) \otimes (0,
\xi_3), \qquad \zeta_i = 0,\dots, \xi_i, \qquad i = 1, 2, 3,
\label{eq:microstructure} \end{equation} with properties \begin{equation}
    \mathbb{M}(\zeta_i, \omega) = \begin{cases} 1 \quad \text{if} \quad \zeta_i
    \in \mathbb{M}^I(\omega) \\ 0 \quad \text{if} \quad \zeta_i \in
\mathbb{M}^M(\omega), \end{cases} \label{eq:microstructure2} \end{equation}
where $\xi_i$ are dimensions and $\zeta_i$ are coordinates inside the unit cell
$\mathbb{M}(\zeta_i, \omega)$, which consists of a matrix phase
$\mathbb{M}^M(\omega)$ and inclusion phases $\mathbb{M}^I(\omega),\quad
\mathbb{M}(\zeta_i, \omega) = \mathbb{M}^M(\omega) \cup \mathbb{M}^I(\omega)$.
The uncertain indicator function 
in 
\autoref{eq:microstructure2} identifies the
different phases at different coordinates $\zeta_i$ 
\ah{(see e.g. \cite{clement2013uncertainty})}. In this work the outer
dimensions $\xi_i$ of the unit cell $\mathbb{M}(\zeta_i, \omega) :=
\mathbb{M}(\omega)$ are deterministic and constant. An uncertain micro
elasticity problem is defined as \begin{align} &1. \quad \nabla \cdot
    \boldsymbol{\sigma}(\omega) = \mathbf{0} \nonumber \\ &2. \quad
    \boldsymbol\sigma(\omega) = \mathbb{C}(\omega) :
    \boldsymbol\varepsilon(\omega) \nonumber \\ &3. \quad
    \boldsymbol\varepsilon(\omega) = \frac{1}{2}(\nabla \otimes
    \mathbf{u}(\omega) +        (\nabla \otimes \mathbf{u}(\omega))^T) \nonumber
    \\ &4. \quad \bar{b}(\omega) \quad \text{on} \quad \partial \mathbb{M},
\label{eq:uequilibrium} \end{align} where $\boldsymbol{\sigma}(\omega)$ is an
uncertain micro stress tensor, $\mathbb{C}(\omega)$ an uncertain micro
elasticity tensor, $\boldsymbol\varepsilon(\omega)$ an uncertain micro strain
tensor and $\mathbf{u}(\omega)$ an uncertain micro displacement tensor.
\rv{Here, Eq. (25.1)
denotes equilibrium conditions, Eq. (25.2) and Hooke's law, Eq. (25.3)
strain-displacement conditions. For the
boundary value problem in \autoref{eq:uequilibrium} to be well posed, uncertain
boundary conditions $\bar{b}(\omega)$ are introduced in Eq. (25.4), where $\partial
\mathbb{M}$ denotes the boundary of the corresponding unit cell
$\mathbb{M}(\omega)$ in \autoref{eq:microstructure}. Possible choices are 
Dirichlet, Neumann or periodic boundary conditions.} 
    
Accounting also for periodicity, the uncertain micro strain
tensor $\boldsymbol\varepsilon(\omega)$ in Eq. (25.3) can be split into an
average deterministic macro strain tensor $\bar{\boldsymbol\varepsilon}$ and an
$\mathbb{M}$-periodic fluctuating uncertain micro strain tensor
$\boldsymbol\varepsilon^* (\omega)$ \begin{equation}
\boldsymbol\varepsilon(\omega) = \bar{\boldsymbol\varepsilon} +
\boldsymbol\varepsilon^* (\omega),\qquad 
\displaystyle\int_{\mathbb{M}} \boldsymbol{\varepsilon}^*(\omega)
\mathrm{d}\boldsymbol{\zeta} =
\boldsymbol{0}, 
\label{eq:BC} \end{equation} 
where the
fluctuating part $\boldsymbol\varepsilon^* (\omega)$ must be compatible
(continuous and single-valued) to an $\mathbb{M}$-periodic displacement field.
The uncertain micro elasticity tensor $\mathbb{C}(\omega)$ in Eq.(25.2) depends
on uncertain material parameters $\boldsymbol{\kappa}(\omega)$ of phases $(i)$
in \autoref{eq:microstructure2}, such that 
\begin{equation}
\rv{\mathbb{C}(\boldsymbol\kappa^{(i)}(\omega)) = K^{(i)}(\omega) \mathbf{1}
\otimes \mathbf{1} + 2G^{(i)}(\omega)
\mathbb{I}^{\text{dev}},\quad i = M, I,}
\label{eq:elten} 
\end{equation} 
with 
\begin{equation} \rv{\boldsymbol\kappa(\omega)
    = \{\kappa_1^{(i)}(\omega), \kappa_2^{(i)}(\omega)\} \equiv
    \{\lambda^{(i)}(\omega), G^{(i)}(\omega)\} \equiv \{\text{E}^{(i)}(\omega),
\nu^{(i)}(\omega)\} \equiv \{\text{K}^{(i)}(\omega), \rv{G^{(i)}(\omega)}\},}
\label{eq:matpar} 
\end{equation} 
where $\lambda^{(i)}(\omega)$ is the uncertain
first Lam\'{e} constant, \rv{$G^{(i)}(\omega)$} the uncertain shear modulus,
$E^{(i)}(\omega)$ the uncertain Young's modulus, $\nu^{(i)}(\omega)$ the
uncertain Poisson's ratio and $K^{(i)}(\omega)$ the uncertain bulk modulus.
Solving the uncertain boundary value problem in \autoref{eq:uequilibrium} on an
uncertain microstructure $\mathbb{M}(\omega)$ in \autoref{eq:microstructure} and
using an average operator on the micro fields, one obtains corresponding
effective macro fields 
\begin{equation} 
    \bar{\boldsymbol\varepsilon} = \langle \boldsymbol\varepsilon(\omega)
    \rangle,\quad \bar{\boldsymbol\sigma}(\omega) = \langle
    \boldsymbol\sigma(\omega) \rangle,\quad 
    \ah{\bar{\boldsymbol\sigma}(\omega) = \bar{\mathbb{C}}(\omega) :
    \bar{\boldsymbol\varepsilon}(\omega)}
    \label{eq:average_field} 
\end{equation}
    where $\bar{\boldsymbol{\sigma}}(\omega)$ denotes an uncertain effective
    macro stress tensor, $\bar{\boldsymbol{\varepsilon}}$ a deterministic
    effective macro strain tensor and $\bar{\mathbb{C}}(\omega)$ an uncertain
    effective macro elasticity tensor. The average operator $\langle \bullet
    \rangle$ in \autoref{eq:average_field} is defined as \begin{equation}
    \langle \bullet \rangle = \displaystyle\frac{1}{\mathbb{M}}
\displaystyle\int_{\mathbb{M}} (\bullet) \mathrm{d}\mathbb{M}.  \end{equation}
The macro and micro stress and strain fields need to satisfy the
\textit{Hill-Mandel condition}: \begin{equation} \langle
    \boldsymbol{\sigma}(\omega) : \boldsymbol{\varepsilon}(\omega) \rangle =
    \langle \boldsymbol{\sigma}(\omega) \rangle : \langle
    \boldsymbol{\varepsilon}(\omega) \rangle.  \label{eq:Hill_Mandel}
\end{equation}
Numerical homogenization can be represented by a model $\mathcal{M}$ in
\autoref{eq:multivariate_random} with random input variable in
$\mathbf{X}(\omega)$ \autoref{eq:input} and random output variable
$\mathbf{Y}(\omega)$ in \autoref{eq:multivariate_random} defined as
\begin{align} 
    &1.\quad \mathbf{X}(\omega) = \{\boldsymbol{\kappa}(\omega),
    \mathbb{M}(\omega), \bar{b}(\omega)\},\nonumber \\ 
    &\rv{2.\quad \mathbf{Y}(\omega) = \bar{\mathbb{C}}(\omega) 
    = \mathcal{M}(\boldsymbol{\kappa}(\omega), \mathbb{M}(\omega), 
\bar{b}(\omega)).}  
    \label{eq:homo_qoi} 
\end{align} 
    The model $\mathcal{M}$ in
    \autoref{eq:homo_qoi} can be realized e.g. by an FFT-based homogenization
    method $\mathcal{M} = \mathcal{M}_{FFT}$, based on
    \cite{vondrejc_fft-based_2014, de_geus_finite_2017, zeman2017finite}.
    \ah{Following \cite{zeman2017finite}, a brief overview of the FFT-based
        homogenization scheme is sketched. For detailed explanations, the reader 
        is
        referred to the above mentioned papers. As a point of departure, the 
        uncertain micro elasticity problem
    is recast into the weak form using test strains $\delta
    \boldsymbol{\varepsilon^*}$, 
    such that}
    \ah{
    \begin{equation}
        \displaystyle\int_{\mathbb{M}} \delta \varepsilon^* :
        \boldsymbol{\sigma}(\boldsymbol{\zeta}, \bar{\boldsymbol\varepsilon} +
    \boldsymbol\varepsilon^* (\omega)) \mathrm{d}\boldsymbol{\zeta} 
    = \displaystyle\int_{\mathbb{M}} \left[\mathbb{G} \star \delta 
    \varepsilon^* \right] 
    : \boldsymbol{\sigma}(\boldsymbol{\zeta}, \bar{\boldsymbol\varepsilon} +
    \boldsymbol\varepsilon^* (\omega)) \mathrm{d}\boldsymbol{\zeta}
    = 0, 
    \end{equation}}\ah{where the compatibility of the test strains is 
        enforced by means of a
    convolution $\star$ with a projection operator to compatible solutions 
$\mathbb{G}$. This projection operator also enforces the zero-mean condition in
\autoref{eq:BC} and is known analytically in Fourier space.
Carrying out discretization by means of trigonometric polynomials and solving 
integrals by
quadrature rules, the following expression in matrix notation holds}
\ah{
\begin{equation}
    \underline{\underline{\mathbb{G}}} : 
    \underline{\underline{\sigma}}(\underline{\underline{\varepsilon}}(\omega))
    = \underline{0}.
    \label{eq:fft}
\end{equation}}\ah{The model $\mathcal{M}_{\text{FFT}}$ is then
established by solving for the
unknown micro strain field $\boldsymbol{\varepsilon}(\omega)$ in 
\autoref{eq:fft} using strain boundary conditions and  projection based 
iterative methods such as e.g.
the conjugate gradient method. After solving \autoref{eq:fft}, the uncertain
effective macro elasticity tensor can be recovered from 
\autoref{eq:average_field}.}
By prescribing the uncertain macro strain tensor $\bar{b}(\omega) =
    \bar{\boldsymbol{\varepsilon}}(\omega)$ from \autoref{eq:uequilibrium}, 
    Eq. (32.2) 
    then
    becomes \begin{equation} \bar{\mathbb{C}}(\omega) \approx
    \mathcal{M}_{{FFT}}(\boldsymbol{\kappa}(\omega), \mathbb{M}(\omega),
    \bar{\boldsymbol{\varepsilon}}(\omega)) =: \bar{\mathbb{C}}_{FFT}(\omega).
\label{eq:fft_model} 
\end{equation}   
    \textit{Remarks:} 
    \begin{enumerate} 
        \item
            Attempts in the literature to use ANN in microstructural
            homogenization were either restricted to two dimensions, uniaxial
            strain or fixed material parameters, using FEM instead of FFT. A
            short overview is given in \autoref{tab:overview}.
    \item For uncertain homogenization calculating the full elasticity tensor
        ${\mathbb{C}}(\omega)$ in \autoref{eq:elten}, three dimensional
        microstructures $\mathbb{M}(\omega)$ in \autoref{eq:microstructure},
        multiaxial strain $\mathbf{\varepsilon}(\omega)$ in
        \autoref{eq:average_field} and varying material parameters
        $\mathbf{\kappa}(\omega)$ in \autoref{eq:matpar} are needed.  
    \item For
        complex microstructures, where meshing for FEM becomes \rv{expensive}, 
        FFT as
        in \autoref{eq:fft} is a mesh free alternative.  
    \rv{\item It has to be carefully distinguished between an untrained ANN and 
        a
    trained ANN. The trained ANN is deterministic, as it's weights are fixed
after optimization. For the rest of the paper, the untrained ANN will be denoted
by $\tilde{\mathcal{M}}_{ANN}^{}$ and the trained ANN by 
$\tilde{\mathcal{M}}_{ANN}^{t}$.}
    \item 
\ah{ 
    There are several choices for boundary conditions to solve the boundary 
    value
    problem in Eq. (25). In this work, the FFT-based Galerkin method of 
    \cite{vondrejc_fft-based_2014, de_geus_finite_2017, zeman2017finite}
    is used, which works directly with strains instead of displacements and 
    enforces compatibility of
the solution by a projection operator to compatible solutions. Therefore, strain
boundary conditions are used.}
    \item The following
        sections propose an algorithm to take these points into account, namely
        by considering three dimensional, uncertain microstructures with
        uncertain material parameters and multiple loading directions.
\end{enumerate}

\begin{table}[htb] \center \begin{tabular}{@{}lllll@{}} \toprule Author &
        Microstructures & Material parameters & Loading & Uncertainty\\ \midrule
        \cite{yang_deep_2018} & three dimensional & fixed & one direction &
        fully deterministic\\ \cite{frankel_predicting_2019} & two dimensional &
        fixed & one direction & fully deterministic\\ \cite{beniwal_deep_2019} &
        two dimensional & fixed & one direction & fully deterministic\\
        \cite{rao_three-dimensional_2020} & three dimensional & fixed & multiple
        directions & uncertain microstructure\\ \midrule Present & three
    dimensional & variable & multiple directions & fully uncertain\\ \bottomrule
\end{tabular} \caption{Overview of similar homogenization methods using ANN from
selected authors.} \label{tab:overview} \end{table}

\section{A deep learning uncertain FFT algorithm}
\label{sec:Algorithmic_Procedure} In a previous work,
\cite{caylak_polymorphic_nodate} investigate an uncertain numerical
homogenization method of long fiber reinforced plastics. Uncertainties of
material parameters and geometry are considered and modeled by multivariate
random variables. The homogenization is carried out by the finite element method
(FEM) utilizing periodic boundary conditions over a meshed representative volume
element. To propagate the input uncertainties and calculate effective uncertain
properties after homogenization, an intrusive Galerkin projection PCE is used.
In this work, an extension towards more complex microstructures is made. In the
following, the framework of the proposed method is presented, followed by
detailed explanations of the implementation.

\subsection{Uncertain homogenization framework} 
\rv{It was shown in e.g. \cite{cruzado2021variational} and
    \cite{VONDREJC2020112585}, that FFT based
    homogenization schemes outperform FEM based full
    field homogenization techniques with respect to CPU time and memory
    requirement. 
FFT based homogenization
methods, as described in \autoref{eq:fft}, are meshless. The discretization
of the geometry is carried out on a regular grid utilizing voxels.}
Homogenization using FFT is described by the model $\mathcal{M}_{FFT}$ in
\autoref{eq:fft_model}, which is used for the \rv{evaluation} in 
\autoref{eq:homo_qoi}.
The goal is to calculate the uncertain effective macro elasticity tensor
$\bar{\mathbb{C}}(\omega)$ in \autoref{eq:average_field}. To calculate central
moments like mean in \autoref{eq:moment1} and variance in \autoref{eq:moment2}
of components of the uncertain effective macro elasticity tensor
$\bar{\mathbb{C}}(\omega)$ in \autoref{eq:average_field}, UQ is needed. In this
work, \rv{a PCE according to \autoref{eq:surrogate_pce} as a surrogate model
$\tilde{\mathcal{M}}_{PCE}$ following \autoref{eq:surrogate1} is used to
approximate \autoref{eq:fft_model}} 
\rv{
\begin{equation} 
        \bar{\mathbb{C}}_{FFT}(\omega) \approx
        \tilde{\mathcal{M}}_{PCE}(\boldsymbol{\kappa}(\omega),
        \mathbb{M}(\omega), \bar{\boldsymbol{\varepsilon}}) =
    \displaystyle\sum_{|\mathbf{i}| \leq n_{PCE}} \hat{\mathbb{C}}_{\mathbf{i}}
\Psi_{\mathbf{i}}(\mathbf{\theta}(\omega)) =: \bar{\mathbb{C}}_{PCE}(\omega),
\label{eq:PCE_C} 
\end{equation}}\rv{where $\mathbf{i}$ is the 
    multi-index 
in \autoref{eq:mulind}. In order to avoid PC arithmetic like in
\cite{caylak_polymorphic_nodate}, which is needed for the intrusive Galerkin
PCE, the pseudospectral approach in \autoref{eq:pseudo} may be used 
to approximate the PC
coefficients $\hat{\mathbb{C}}_{\mathbf{i}}$ in \autoref{eq:PCE_C}}
\rv{
\begin{equation} 
    \hat{\mathbb{C}}_{\mathbf{i}} \approx \hat{\mathbb{C}}_{\mathbf{i}}^{FFT} =
    \frac{1}{\gamma_{\mathbf{i}}}
\displaystyle\sum_{j=1}^{n_w}{\mathcal{M}_{{FFT}}}(\Theta^{(j)})
\Psi_{\mathbf{i}}(\Theta^{(j)})w(\Theta^{(j)}),\quad
j = 1, \dots, n_w. 
\label{eq:coeff} 
\end{equation}}\rv{Here, $\Theta^{(j)}$ and $w(\Theta^{(j)})$ are the nodes and
weights of the cubature rule in \autoref{eq:dataset_cub}, which can be gathered
in the dataset $\mathbb{D}_{cub} = \mathbb{D}_{FFT}$ from
\autoref{eq:dataset_cub} in single index notation. To this end, $n_q$
deterministic solutions of the computational model $\mathcal{M}_{{FFT}}$, as
defined in \autoref{eq:no_solutions}, are needed in \autoref{eq:coeff}}
\begin{equation} 
    \mathbb{D}_{FFT} = \left\{(\mathbf{x}_k = \Theta_k, \dots,
\mathbf{x}_{n_q} = \Theta_{n_q}), (\mathcal{M}_{FFT}(\Theta_k), \dots,
\mathcal{M}_{FFT}(\Theta_{n_q})), \quad k=1,\dots,n_{q}\right\}.
\label{eq:dataset_cub_C} 
\end{equation}
In \autoref{eq:dataset_cub_C},
$\mathbf{x}_k = \{\boldsymbol{\kappa}(\omega), \mathbb{M}(\omega),
\bar{\boldsymbol{\varepsilon}}\}$ are multivariate realizations of the random
input variable $\mathbf{X}(\omega)$ in \autoref{eq:homo_qoi}.
\rv{Finally 
    the pseudospectral PCE follows from 
    \autoref{eq:PCE_C} and 
\autoref{eq:coeff} as
\begin{equation}
    \bar{\mathbb{C}}_{PCE}(\omega) \approx \displaystyle\sum_{|\mathbf{i}| \leq
    n_{PCE}} {\hat{\mathbb{C}}^{FFT}_{\mathbf{i}}}
\Psi_{\mathbf{i}}(\mathbf{\theta}(\omega)).  
\label{eq:main_problem}
\end{equation}}

For large unit
cells $\mathbb{M}(\omega)$ in \autoref{eq:microstructure}, the solution of the
problem formulated in \autoref{eq:main_problem} becomes computational
challenging due to the large number of evaluations $n_q$ of the deterministic
model $\mathcal{M}_{FFT}$, which are needed for the PCE as described in
\autoref{eq:no_solutions}. Instead, an ANN 
$\tilde{\mathcal{M}}_{ANN}$ as
defined in \autoref{eq:nn} can be trained to learn the deterministic model
$\mathcal{M}_{FFT}$, which is faster to evaluate than the original model,
because only simple matrix multiplications in FFNN \autoref{eq:FFNN} and CNN
\autoref{eq:CNN} are needed. \rv{This approximation can be formulated as
\begin{equation} 
    \mathcal{M}_{\text{FFT}} \approx
    \tilde{\mathcal{M}}_{{ANN}}^t.
\label{eq:surr_ann_fft} 
\end{equation} 
\autoref{eq:coeff} is replaced by 
\begin{equation} 
 \hat{\mathbb{C}}_{\mathbf{i}} \approx \hat{\mathbb{C}}^{ANN}_{\mathbf{i}} = 
\frac{1}{\gamma_{\mathbf{i}}}
            \displaystyle\sum_{j=1}^{n_w}\tilde{\mathcal{M}}_{{ANN}}^t
            (\Theta^{(j)})
            \Psi_{\mathbf{i}}(\Theta^{(j)})w(\Theta^{(j)}).
\label{eq:coeff2} 
\end{equation}
For the approximation
\autoref{eq:surr_ann_fft} to hold, an ANN $\tilde{\mathcal{M}}_{ANN}$ needs to
be trained on a dataset $\mathbb{D} = \mathbb{D}_{ANN}$ in \autoref{eq:dataset}
defined as 
\begin{equation} 
    \mathbb{D}_{ANN} = \left\{(\mathbf{x}_k = \left\{
                \boldsymbol{\kappa}_k, \boldsymbol{\mathbb{M}}_k,
    \boldsymbol{\bar{\varepsilon}}_k \right\}), (\mathbf{y}_k =
\boldsymbol{\bar{\sigma}} =       \mathcal{M}_{FFT}(\mathbf{{x}}_k),\quad k = 1,
\dots, n_s\right\}, 
\label{eq:dataset_ann} 
\end{equation} where $\mathbf{x}_k$ 
\rv{need to sample the physically admissible support} from $\mathbf{X}(\omega)$ 
in \ah{\autoref{eq:homo_qoi}}. 
Finally \autoref{eq:main_problem} is approximated as 
\begin{equation}
    \bar{\mathbb{C}}_{PCE}(\omega) \approx
    \displaystyle\sum_{|\mathbf{i}| \leq \rv{n_{PCE}}}
    {\hat{\mathbb{C}}^{ANN}_
    {\mathbf{i}}}
    \Psi_{\mathbf{i}}(\mathbf{\theta}(\omega)) =:
    {\bar{\mathbb{C}}}_{ANN}(\omega).  \label{eq:main_problem2}
\end{equation}}This is the final formulation for the proposed UQ FFT model using
an ANN $\tilde{\mathcal{M}}_{ANN}$. Once
trained, $\tilde{\mathcal{M}}_{ANN}^t$ provides the 
\ah{\textit{deterministic solutions} for
the macro elasticity tensor ${\bar{\mathbb{C}}}_{ANN}$ in
\autoref{eq:main_problem2}.} From the surrogate $\tilde{\mathcal{M}}_{PCE}$ in
\autoref{eq:main_problem2}, central moments, mean $\boldsymbol{\mu}$ 
\autoref{eq:moment1} and
variance $\boldsymbol{\sigma}^2$ 
\autoref{eq:moment2}, can be calculated. In the following
section, details of the implementation of \autoref{eq:main_problem2} are
provided.

 \subsection{Numerical implementation} \subsubsection{Algorithm overview} To
realize \autoref{eq:main_problem2}, \autoref{algo:general_procedure} must be
implemented. Its task is to generate a training set $\mathbb{D}_{ANN}$
\autoref{eq:dataset_ann} for training an ANN $\tilde{\mathcal{M}}_{{ANN}}$ to
learn FFT homogenization \autoref{eq:fft_model}, which is then used for
uncertainty quantification of a homogenized  elasticity tensor
${\bar{\mathbb{C}}}_{ANN}$ in \autoref{eq:main_problem2}. 

\ah{\textit{Remark 1}: It has to be pointed out, that the ANN 
        $\tilde{\mathcal{M}}_{{ANN}}^t$ is a purely \textit{deterministic}
        surrogate to the deterministic FFT solver from \autoref{eq:fft}. After
        training, the weights of the ANN in \autoref{eq:FFNN} and
        \autoref{eq:CNN} are fixed, thus producing
deterministic outputs for given deterministic inputs.}

The single
steps, namely data generation in \autoref{sec:Data_Creation}, \rv{ANN} 
training in
\autoref{sec:ANN_Creation} and UQ in \autoref{sec:UQ}, are further described in
detail.
\algsetup{indent=2em} 
\begin{algorithm}[htb] 
    \renewcommand{\thealgorithm}{0}
    \caption{Overall procedure for UQ with PCE and ANN} 
    \label{algo:general_procedure} 
    \begin{algorithmic}[0]
        \STATE \rv{\textbf{Algorithm 1: Data Generation
        for Deep Learning using FFT}} 
        \STATE \qquad 
\rv{ 
        \textbf{Input:} $\mathbf{X}^U(\omega) = \{\boldsymbol{\kappa}^U(\omega),
        \mathbb{M}^U(\omega), \bar{\boldsymbol{\varepsilon}}\} \sim
        \boldsymbol{\mathcal{U}}(\underbar{x}, \bar{x})$ in
\autoref{eq:input2}}
        \STATE \qquad
        \rv{Homogenization: $\mathbf{{y}}_k = \boldsymbol{\bar{\sigma}} =
        \mathcal{M}_{FFT}(\mathbf{{x}}_k)$ in \autoref{eq:main_out}}
        \STATE \qquad
        \rv{Output:  
            training dataset $\boldsymbol{\mathbb{D}}_{ANN} =
            \left\{(\mathbf{{x}}_k),       (\mathbf{{y}}_k), \;
            k=1,\dots,n_s\right\}, \; 
            \displaystyle\frac{n_s}{6} \in \mathbb{N}^+$
        in \autoref{eq:dataset_ann}}
        \STATE \qquad
        \STATE \rv{\textbf{Algorithm 2: ANN Design and Training}}
        \STATE \qquad
        Input: $\rv{\boldsymbol{\mathbb{D}}_{ANN},
        \tilde{\mathcal{M}}_{ANN}^t}$ in
        \autoref{eq:dataset_ann} and \autoref{eq:nn} \STATE \qquad
        Training: $\underset{\mathbf{W}, \mathbf{K}}{\arg \min
        }\{\left\|\mathbf{y}_{k}-\tilde{\mathbf{y}}_{k}(\mathbf{W},
            \mathbf{K})\right\|_{2}+\underbrace{\lambda_{L 2}\|\mathbf{W},
        \mathbf{K}\|_{2}^{2}}_{\mathcal{R}(\mathbf{W}, \mathbf{K})}\}$ in
        \autoref{eq:optimization} \STATE \qquad 
        Output: 
        \rv{$\mathbf{W}, \mathbf{K}, \tilde{\mathcal{M}}_{{ANN}}^t$ in 
        \autoref{eq:surr_ann_fft}}
        \STATE 
        \STATE \rv{\textbf{Algorithm 3: UQ using PCE and ANN trained on FFT}}
        \STATE \qquad Input: $\tilde{\mathcal{M}}_{ANN}^t, 
        \mathbf{X}(\omega) =
        \{\boldsymbol{\kappa}(\omega), \mathbb{M}(\omega),
        \hat{\boldsymbol{\varepsilon}}\} \sim
        \boldsymbol{\mathcal{N}}(\boldsymbol{\mu}, \mathbf{\Sigma})$ in
        \autoref{eq:surr_ann_fft} and \autoref{eq:homo_qoi} \STATE \qquad 
        Output:
        $\rv{{\bar{\mathbb{C}}}_{ANN} =
        \tilde{\mathcal{M}}_{\text{PCE}}(\tilde{\mathcal{M}}_{{ANN}}^t
            (\boldsymbol{\kappa}(\omega),\boldsymbol{\mathbb{M}}(\omega),
    \boldsymbol{\bar{\varepsilon}}_k))}$ in \autoref{eq:main_problem2}
    \end{algorithmic} 
\end{algorithm}

\subsubsection{Data generation for deep learning using FFT}
\label{sec:Data_Creation} 
In step 1 of Algorithm \ref{algo:general_procedure}, 
a dataset $\mathbb{D}_{ANN}$
from \autoref{eq:dataset_ann} is created with Algorithm
\ref{algo:data_creation}. \rv{The inputs to the dataset $\mathbb{D}_{ANN}$ are
    realizations $\mathbf{{x}}_k$ of an input random variable denoted by 
    $\mathbf{X}^U(\omega)$ from
    \autoref{eq:input}, such that
    \begin{equation} 
        \mathbf{X}^U(\omega) = \mathbf{{x}}_k = \left\{ \boldsymbol{\kappa}_k,
        \boldsymbol{\mathbb{M}}_k, \boldsymbol{\bar{\varepsilon}}_k \right\} \in
        \mathcal{D}_{\boldsymbol{X}}^U, \quad k = 1, \dots, n_s,
        \label{eq:input2} 
    \end{equation} 
    where $\mathbb{M}_k$ are microstructures as 
    described in \autoref{eq:microstructure},
    $\boldsymbol{\kappa}_k$ are material parameters defined in 
    \autoref{eq:matpar}, $\bar{\boldsymbol{\varepsilon}}_k$
are macro strains from \autoref{eq:average_field} and 
$\mathcal{D}_{\boldsymbol{X}}^U$ denotes the sample space of the training input
variables.} 
The total number of samples of the dataset
$\mathbb{D}_{ANN}$ is denoted by $n_s$, consistent with \autoref{eq:dataset}. 

\rv{Generally, the only influence on the outcome of
    the ANN $\tilde{\mathcal{M}}_{ANN}^t$ in \autoref{eq:surr_ann_fft} 
    is the training process including the data provided during
    training. The choice of inputs in \autoref{eq:input2} 
    for the training dataset $\mathbb{D}_{ANN}$
    from \autoref{eq:dataset_ann} is therefore very important. Clustering around
    specific values of e.g. the fiber volume fraction $c_{f,k}$ or the material
    parameters $\kappa_k$ in the dataset could lead to a bias towards these
    values.  Having in mind the goal to establish a deterministic surrogate to
    replace the FFT solver from \autoref{eq:fft}, the inputs from
    \autoref{eq:input2} to the dataset $\mathbb{D}_{ANN}$ should fill their
    sample space ${\mathcal{D}}_{\boldsymbol{X}}^U$ from \autoref{eq:input2}
    uniformly. Therefore, the inputs $x_k$ from \autoref{eq:input2} are drawn
    from uniform 
    distributions $\boldsymbol{X}^{U}(\omega) \sim
    \boldsymbol{\mathcal{U}}(\underbar{x}, \bar{x})$.
    Additionally, physically inadmissible values must be avoided. 
    Therefore, the sample
    space ${\mathcal{D}}_{\boldsymbol{X}}^U$ from \autoref{eq:input2} needs to
    be restricted. 
    In implementation practise, these values
    are drawn from uniform distributions, whereas the lower and upper bounds
    $\underbar{x}$ and $\bar{x}$, respectively, have
    to be chosen in accordance with physical restrictions. Therefore
    the general expression of ${\mathcal{D}}_{\boldsymbol{X}} \subset
    \mathbb{R}$ is specified  
    from \autoref{eq:input} to \autoref{eq:input2} as
    \begin{equation}
        \mathcal{D}_{\boldsymbol{X}}^U = 
        \begin{cases} 
            \mathcal{D}_{\boldsymbol{X}}^{{\kappa}_i} =
            \{{\kappa}_i: [\underline{\kappa}_i \leq
                    {\kappa}_i 
            \leq \bar{\kappa}_i]\}, \; i = 1, \ldots, n_p   \\
            \mathcal{D}_{\boldsymbol{X}}^{c_f} = \{c_f: [0 \leq c_f \leq 1]\}.
        \end{cases}
        \label{eq:phyconst}
    \end{equation}
    Here, ${\kappa}_i$ denote the single material parameters from
    \autoref{eq:matpar}. The total number of material parameters is denoted by
    $n_p$. The upper and lower bounds, $\underline{\kappa}_i$ and
    $\bar{\kappa}_i$, respectively, are chosen according to the respective
    admissible
    range of the material parameter, e.g. 
$[\underline{\nu} = 0 \leq \nu \leq \bar{\nu} = 0.5]$ for Poisson's ratio,
which is the range for typical engineering materials, including those considered
in this work.}

The microstructure $\mathbb{M}_k$ from \autoref{eq:input2} is 
implemented as a
three dimensional voxel array, where the entries are binary, such that $0$ is
used for matrix and $1$ for inclusion material. This is mathematically described
by the indicator function $\mathbb{M}(\zeta_i, \omega)$ in
\autoref{eq:microstructure2}. \rv{The microstructure $\mathbb{M}$ depends on
    the fiber volume fraction $c_f$ as 
    \begin{equation}
        \mathbb{M}_k = \mathbb{M}_k(c_{f,k}),
    \label{eq:microstructure_uniform} \end{equation}
    where $c_{f,k}$ is specific to
each microstructure for each $k$ in \autoref{eq:input2}.} In this work,
two kinds of microstructures are considered, namely single long fiber
cylindrical inclusions and multiple spherical inclusions. The latter are
generated by the \textit{random sequential adsorption method}, as described in
\cite{bargmann2018generation}. 


Furthermore, shear and bulk modulus, $G$ and
\ah{$K$}
as described in \autoref{eq:matpar}, respectively, are used as material
parameters in the implementation. 
They are better suited for ANN, because their ranges are of the same
magnitude compared to the combination Young's modulus $E$ and Poisson's ratio
$\nu$ in \autoref{eq:matpar}, which leads to smoother gradient updates in the
optimization defined in \autoref{eq:update}. Other material parameters are
converted internally. 

The output $\mathbf{y}_k$ for the
error measure $\mathcal{E}$ in \autoref{eq:optimization} is calculated by FFT
from \autoref{eq:fft}. \rv{In this work, $\mathbf{y}_k$ is the effective 
    macro stress 
    $\mathbf{y}_k = \bar{\boldsymbol{{\sigma}}}_k$ in 
    \autoref{eq:average_field} corresponding 
    to the macro strain $\bar{\boldsymbol{\varepsilon}}_k$ in
    \autoref{eq:average_field} and \autoref{eq:input2}, such that
    \begin{equation} \mathbf{y}_k
        = \mathcal{M}_{\text{FFT}}({\mathbf{x}}_k) =
        \bar{\boldsymbol{{\sigma}}}_k,\qquad n = 1, \dots, n_s. \label{eq:main_out}
    \end{equation} 
    The effective elasticity tensor $\bar{\mathbb{C}}_k$ 
    in \autoref{eq:average_field} can be reconstructed using Voigt notation such
    that 
    \begin{equation}
        \bar{{\mathbb{C}}} = \left[\bar{\underline{\underline{\mathbb{C}}}}\right] =
        [\bar{\underline{{\sigma}}}_1, \bar{\underline{{\sigma}}}_2,
            \bar{\underline{{\sigma}}}_3, \bar{\underline{{\sigma}}}_4,
        \bar{\underline{{\sigma}}}_5, \bar{\underline{{\sigma}}}_6],\qquad
        \bar{\underline{{\sigma}}}_i = \langle \underline{\underline{\mathbb{C}}} \;
        \underline{\bar{\boldsymbol{\varepsilon}}}_i \rangle, \; i = 1,\cdots,6.
    \label{eq:homogenization} \end{equation} 
    \begin{equation*}
        \underline{\bar{\boldsymbol{\varepsilon}}}_i(j) = \begin{cases} 1 \quad
            \text{for} \quad i = j, \; j = 1,\cdots,6 \\ 0 \quad \text{else},  
        \end{cases}
    \end{equation*} 
    e.g. for $i=1, \quad \underline{\bar{\boldsymbol{\varepsilon}}}_i=1 
    = [1 \; 0 \; 0 \; 0 \; 0 \; 0]^T.$
    The different strain states 
    $\underline{\bar{\boldsymbol{\varepsilon}}}_i$ for
    \autoref{eq:homogenization} need to be equally represented in the dataset
    \autoref{eq:dataset_cub_C}. If one strain state would be
    underrepresented, the predictive capability for that specific state
    $\underline{\bar{\boldsymbol{\varepsilon}}}_i$ would be poor. Therefore, the
    total number of samples $n_s$ must be a multiple of 6, which implies $1/6 \; n_s
    \in \mathbb{N}^+$ or \begin{equation} n_{\boldsymbol{\bar{
        \underline\varepsilon}}_i} = n_{\boldsymbol{\bar{
        \underline\varepsilon}}_j}, \; i,j = 1,\cdots,6, \label{eq:modulo}
    \end{equation} where $n_{\boldsymbol{\bar{ \underline\varepsilon}}_i}$ denotes
    the number of samples with strain state $i$. In practise, a compromise between 
    the number
of samples $n_s$ and the training time has to be made.}
In the examples in
\autoref{sec:Examples}, $n_s = 9000$ for Example 1, $n_s = 13800$ for Example 2
and $n_s = 9000 + 13800 = 22800$ for Example 3. The whole process of data
generation can be carried out on a workstation with GPU acceleration or on a
cluster, because every sample generation is pleasingly parallel.

\ah{\textit{Remark 2:} An alternative approach to \autoref{eq:homogenization} 
    would be to apply all six strain
    states directly. This however would lead to less
    flexibility of the obtained ANN. The proposed approach in 
    \autoref{eq:homogenization} using single strain
    states $\underline{\bar{\boldsymbol{\varepsilon}}}_i$ 
    allows to consider different stochastic properties for different strain
    states in Algorithm 4. This is the case, if e.g. material parameters are
    obtained from experimental data, which show different deviations for different
loading directions.}
\algsetup{indent=2em} \begin{algorithm}[htb] 
    \renewcommand{\thealgorithm}{1}
    \caption{\textbf{Data
    Generation for Deep Learning using FFT}} \label{algo:data_creation}
    \begin{algorithmic}[0] \STATE 
\rv{ 
        \textbf{Input:} $\mathbf{X}^U(\omega) = \{\boldsymbol{\kappa}^U(\omega),
        \mathbb{M}^U(\omega), \bar{\boldsymbol{\varepsilon}}\} \sim
        \boldsymbol{\mathcal{U}}(\underbar{x}, \bar{x})$ in
\autoref{eq:input2}}
        \STATE
        \textbf{Output:} ${\mathbb{D}}_{ANN} = \left\{({\mathbf{x}}_k),
        ({\mathbf{y}}_k), \; k=1,\dots,n_s\right\}, \;
        \displaystyle\frac{n_s}{6} \in \mathbb{N}^+$ \STATE \hrulefill
        \STATE $n_{\boldsymbol{\bar{ \underline\varepsilon}}_i} = n_s \; / \; 6$
        \STATE $k = 1$ \FOR{$i=1:6$} \STATE
        $\bar{\underline{\boldsymbol{\varepsilon}}} = [0, 0, 0, 0, 0, 0]$ \STATE
        $\bar{\underline{\boldsymbol{\varepsilon}}}[i] = 1$
        \FOR{$q=1:n_{\boldsymbol{\bar{ \underline\varepsilon}}_i}$} 
        \STATE \rv{\text{Sample} $\boldsymbol{\kappa}_k 
        \in \mathcal{D}_{\boldsymbol{X}}$} \STATE
        \rv{Sample $c_f \in \mathcal{D}_{\boldsymbol{X}}$ and generate 
            $\boldsymbol{\mathbb{M}}_k(c_f)$ 
        }
        \label{algo:data_creation_micro}
        \STATE \rv{$\boldsymbol{x}_k = 
            \{\boldsymbol{\kappa}_k, \boldsymbol{\mathbb{M}}_k,
        \boldsymbol{\bar{\varepsilon}}_k\}$}
        \STATE \text{Solve}
        $\mathbb{G} : \boldsymbol{\sigma} = \mathbf{0}$ by FFT homogenization
        \autoref{eq:fft} \label{algo:data_creation_fft} \STATE
        $\bar{{\boldsymbol{\sigma}}} = \langle\boldsymbol{{\sigma}}\rangle
        \xrightarrow{\text{Voigt}} \bar{\underline{\boldsymbol{\sigma}}}$ \STATE
        ${\mathbf{y}}_k = \bar{\underline{\boldsymbol{\sigma}}}$ 
    Dataset $\mathbb{D}_{ANN}$ \STATE $k = k + 1$ \ENDFOR \ENDFOR \STATE
(Shuffle $\boldsymbol{\mathbb{D}}_{ANN}$ row-wise) \end{algorithmic}
\end{algorithm}

\subsubsection{ANN design and training} \label{sec:ANN_Creation} In 
Algorithm \ref{algo:data_creation}, an ANN $\tilde{\mathcal{M}}_{ANN}$
defined in \autoref{eq:nn} is created and trained on a dataset
$\boldsymbol{\mathbb{D}}_{ANN}$ from \autoref{eq:dataset_ann} as described in
Algorithm \ref{algo:training}. This part of the implementation is carried out
with the help of \textit{Tensorflow} \cite{tensorflow2015-whitepaper}. In the
following, the single steps of Algorithm \ref{algo:training} are outlined in
detail.

\algsetup{indent=2em} \begin{algorithm}[htb] 
    \renewcommand{\thealgorithm}{2}
    \caption{\textbf{ANN Design and
    Training}} \label{algo:training} \begin{algorithmic}[0] \STATE
        \textbf{Input:} ${\mathbb{D}}_{ANN} = \left\{({\mathbf{x}}_k),
        ({\mathbf{y}}_k), \; k=1,\dots,n_s\right\}, \;
        \displaystyle\frac{n_s}{6} \in \mathbb{N}^+$ \STATE \textbf{Output:}
        $\rv{\tilde{\mathcal{M}}_{ANN}^t}$ \STATE \hrulefill \STATE
        \textbf{(i)\hspace{1em} Topology set up} \STATE
        \textbf{(ii)\hspace{0.7em} Hyperparameter selection} \STATE
    \textbf{(iii)\hspace{0.75em}Training} \end{algorithmic}
\end{algorithm}

\begin{figure}[htb] \centering
    \includegraphics[width=1\textwidth]{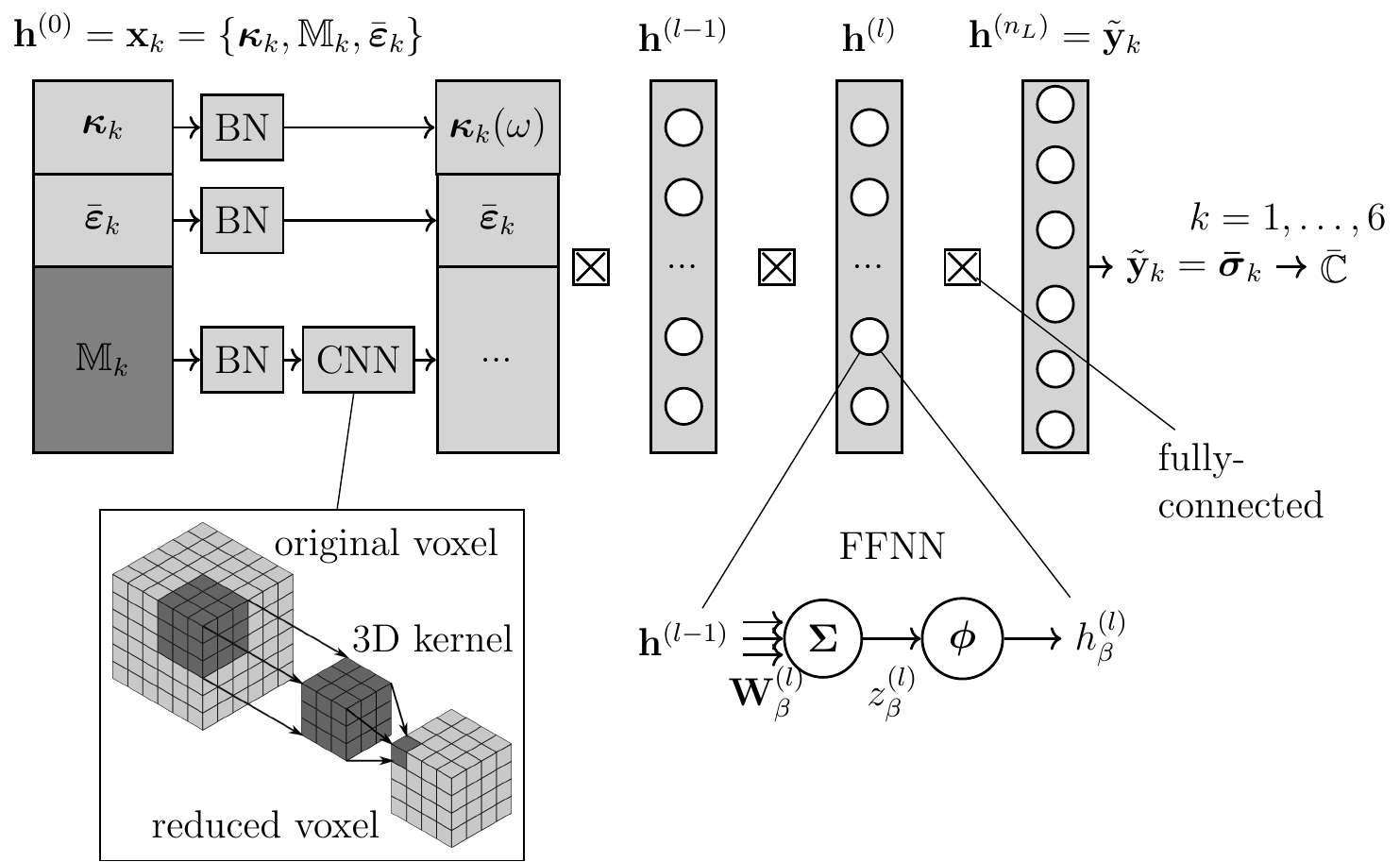} \caption{Proposed
        topology of ANN in \autoref{eq:nn} with $n_L$ layers $\mathbf{h}^{(l)}$ and
        $n_u$ neural units $h_{\beta}^{(l)}$. An ANN with input ${\mathbf{x}}_k =
        \{\boldsymbol{\kappa}(\omega), \mathbb{M}(\omega),
        \bar{\boldsymbol{\varepsilon}}\}$, where $\boldsymbol{\kappa}(\omega)$ is a
        material parameter vector, $\bar{\boldsymbol{\varepsilon}}_k$ a macro strain
        vector and $\mathbb{M}(\omega)$ consist of 3D microstructural voxel data, see
        \autoref{eq:fft_model}. ``BN'' denotes a batch normalization layer, where the
        input is standardized, as described in Section \ref{sec:Topology}.
        $\mathbb{M}(\omega)$ is processed by a CNN in \autoref{eq:CNN} and after
        concatenation with $\boldsymbol{\kappa}(\omega)$ and
        $\bar{\boldsymbol{\varepsilon}}$ feed into an FFNN in \autoref{eq:FFNN}. The
        resulting output is the predicted macroscopic stress tensor
        $\boldsymbol{\tilde{\bar{\sigma}}}_k$ \autoref{eq:main_out}. From six strain
        states $k$, the macroscopic effective elasticity tensor
        ${\bar{\boldsymbol{\mathbb{C}}}}$ can be reconstructed as defined in
\autoref{eq:homogenization}.} \label{fig:ANN_topo} \end{figure}

~\\ \textit{(i) Topology set up} \label{sec:Topology} ~\\ In this work, the ANN
in \autoref{eq:main_problem2} consists of three inputs, corresponding to the
input in \autoref{eq:input2} illustrated in \autoref{fig:ANN}. The first input
consists of the \rv{material parameters $\boldsymbol{\kappa}_k$ in
    \autoref{eq:matpar} and the macro strain
    $\bar{\boldsymbol{\varepsilon}}_k$ in \autoref{eq:average_field}, such that
    $\mathbf{x}_{k,1} = \{ \boldsymbol{\kappa}_k,
    \bar{\boldsymbol{\varepsilon}}_k\}$ in \autoref{fig:ANN_topo}. The second input
    corresponds to the microstructure such that $\mathbf{x}_{k,2} =
\mathbb{M}_k$ from \autoref{eq:microstructure}.} The input in
\autoref{eq:input2} is standardized by a batch normalization layer, denoted by
``BN'' in \autoref{fig:ANN_topo}. This leads to improved training performance
\cite{DBLP:journals/corr/IoffeS15}. The microstructure \rv{$\mathbb{M}$} in
\autoref{eq:microstructure} is processed with a CNN in \autoref{eq:CNN} for
dimension reduction. The architecture of the CNN is either \textit{AlexNet}
\cite{krizhevsky2012imagenet}, such as in \cite{yang_deep_2018} and
\cite{frankel_predicting_2019}, or  \textit{DenseNet} \cite{huang2017densely},
depending on the complexity of the underlying problem. The macro strain vector
$\bar{\boldsymbol{\varepsilon}}_k$ in \autoref{eq:homogenization} and material
parameter vector \rv{$\boldsymbol{\kappa}_k$} in \autoref{eq:matpar} are
concatenated with the reduced output of the CNN in \autoref{fig:ANN_topo}. Then,
multiple FFNN as defined in \autoref{eq:FFNN} follow, each utilizing a rectified
linear unit (ReLU) activation described in \autoref{eq:activation}, L-2
regularization $\mathcal{R}(\mathbf{W}, \mathbf{K})$ in
\autoref{eq:optimization} and dropout as defined in \cite{goodfellow_deep_2016}.
Finally, the output $\tilde{\mathbf{y}}_k$ in \autoref{fig:ANN} is the
macroscopic stress in \autoref{eq:homogenization}, such that
$\tilde{\mathbf{y}}_k = \bar{\mathbf{\sigma}}_k$, from which the macroscopic
elasticity tensor can be reconstructed.

~\\~\\ \textit{(ii) Hyperparameter selection} \label{sec:Hyperparameter} ~\\
Hyperparameters are essential for model performance, i.e. achieving an error in
\autoref{eq:optimization}. In this work, hyperparameters of the ANN in
\autoref{eq:nn} are adjusted by a \textit{random search} algorithm
\cite{bergstra_random_nodate}. A summary of hyperparameters considered in this
work is given in \autoref{tab:hp}. Hyperparameter tuning is carried out on a
separate dataset $\boldsymbol{\mathbb{D}}_{\text{HP}}$, with similar definition
as in \autoref{eq:dataset}, to not overfit to the original dataset
$\boldsymbol{\mathbb{D}}_{\text{ANN}}$.

\begin{table}[htb] \center \begin{tabular}{@{}cll@{}} \toprule Symbol &
        Description & Reference \\ \midrule $\alpha$ & learning rate &
        \autoref{eq:update} \\ $\beta$ &  dropout rate &
        \cite{goodfellow_deep_2016} \\ $\lambda_{L2}$  & L2 parameter &
        \autoref{eq:optimization} \\ $n_u$ &  no. of units (FFNN) &
        \autoref{eq:FFNN} \\ $n_F$ & no. of filters (CNN) &\autoref{eq:CNN} \\ $n_L$
    & no. of layers (FFNN) &  \autoref{eq:nn} \\ \bottomrule \end{tabular}
    \caption{Hyperparameters of ANN in \autoref{eq:nn} considered in this work.}
\label{tab:hp} \end{table}

~\\~\\ \textit{(iii) Training} \label{sec:Training} ~\\ After providing the
dataset $\mathbb{D}_{ANN}$ in Algorithm \ref{algo:data_creation} as well as the
topology and hyperparameters, the ANN $\tilde{\mathcal{M}}_{ANN}$ is trained by
updating its weights \autoref{eq:optimization} with respect to the dataset
$\mathbb{D}_{ANN}$ in \autoref{eq:dataset_ann}. For stochastic gradient descent
used to minimize \autoref{eq:update}, the ADAM optimizer \cite{kingma2014adam}
in its       AMSGrad variant \cite{reddi2019convergence} is used. The weights
$\mathbf{W}$ in \autoref{eq:FFNN} and kernels $\mathbf{K}$ in \autoref{eq:CNN}
are initialized via Glorot Xavier initialization
\cite{glorot2010understanding}. Early stopping and learning rate decay, as
described in \cite{geron2019hands}, are used during optimization
\autoref{eq:optimization}.

\subsubsection{UQ using pseudospectral PCE and ANN trained on FFT}
\label{sec:UQ} In Algorithm \ref{algo:uq}, UQ of the
uncertain effective macro elasticity tensor $\bar{\mathbb{C}}(\omega)$ in
\autoref{eq:main_problem2} using the trained ANN
$\rv{\tilde{\mathcal{M}}_{ANN}^t}$
from \autoref{eq:surr_ann_fft} is carried out. For the implementation of
pseudospectral PCE as defined in \autoref{eq:surrogate_pce}, \textit{ChaosPy}
\cite{feinberg_chaospy:_2015} is used. In the following, the single steps of
Algorithm \ref{algo:uq} are outlined in detail.

\algsetup{indent=2em} \begin{algorithm}[htb] 
    \renewcommand{\thealgorithm}{3}
    \caption{\textbf{: UQ using PCE and
    ANN trained on FFT}} \label{algo:uq} \begin{algorithmic}[0] \STATE
        \textbf{Input:} $\mathbf{X}(\omega) = \{\boldsymbol{\kappa}(\omega),
        \mathbb{M}(\omega), \bar{\boldsymbol{\varepsilon}}\} \sim
        \boldsymbol{\mathcal{N}}(\boldsymbol{\mu}, \mathbf{\Sigma})$ in
        \autoref{eq:homo_qoi} \STATE \textbf{Output:}
        $\boldsymbol{\mu}_{\mathbf{Y}}, \boldsymbol{\sigma}_{\mathbf{Y}}^2$
        \STATE \hrulefill \STATE \STATE \textbf{(i) Cubature rule:}\\\qquad
        Obtain cubature nodes ${\Theta}^{(j)}$ and weights $w({\Theta}^{(j)})$
        from Gaussian cubature rule w.r.t. $\mathbf{X}(\omega)$ \STATE \STATE
        \textbf{(ii) Deterministic solutions:}\\\qquad Calculate deterministic
        solutions at nodes $\rv{\mathcal{M}_{\text{ANN}}^t({\Theta}^{(j)})}$ in
        \autoref{eq:dataset_cub} \STATE \label{eq:det_sol} \STATE \textbf{(iii)
        Orthonormal polynomials:}\\\qquad Generate Hermite polynomial
        ${\Psi}_{\mathbf{i}}$ in \autoref{eq:surrogate_pce} \STATE \STATE
        \textbf{(iv) PC coefficients:}\\\qquad Calculate PC coefficients
        $\rv{\hat{\mathbb{C}}_{\mathbf{i}}^{ANN}} =
        \displaystyle\frac{1}{\gamma_{\mathbf{i}}}
        \displaystyle\sum_{j=1}^{n_w}
        \rv{\mathcal{M}_{\text{ANN}}^t}(\Theta^{(j)})
        \Psi_{\mathbf{i}}(\Theta^{(j)})w(\Theta^{(j)})$
        in \autoref{eq:coeff2} \STATE \STATE \textbf{(v) Statistics:}\\\qquad
        Calculate statistics from uncertain effective elasticity tensor
        \\\qquad$\rv{{\bar{\mathbb{C}}}_{PCE} \approx
            \rv{{\bar{\mathbb{C}}}_{ANN}}(\mathbf{X}(\omega)) =
            \displaystyle\sum_{|\mathbf{i}| \leq n_{PCE}}
            \hat{\mathbb{C}}_{\mathbf{i}}^{ANN}
        \Psi_{\mathbf{i}}(\mathbf{\theta}(\omega))}$ in
\autoref{eq:main_problem2} \end{algorithmic} \end{algorithm}
~\\~\\ 
\textit{Multivariate random input variable}
\label{sec:Sample_Distribution} ~\\ The random variables
$\boldsymbol{\kappa}(\omega)$ in \autoref{eq:matpar} and $\mathbb{M}(\omega)$ in
\autoref{eq:microstructure} from the multivariate random input variable
$\mathbf{X}(\omega)$ as defined in \autoref{eq:fft_model} are normally
distributed, as seen in the input of Algorithm (\ref{algo:uq}). If multiple
material parameters $\boldsymbol{\kappa}(\omega)$ in \autoref{eq:matpar} are
considered, e.g. linear elastic parameters for different constituents as defined
in \autoref{eq:matpar}, each individual parameter is a normally distributed
univariate random variable. For each univariate random variable of
$\mathbf{X}(\omega)$, the mean $\boldsymbol{\mu}$ in \autoref{eq:mean} and
standard deviation $\boldsymbol{\sigma}$ in \autoref{eq:deviation} must be
provided. \textit{ChaosPy} \cite{feinberg_chaospy:_2015} then automatically
ensembles the corresponding multivariate expectation vector $\boldsymbol{\mu}$,
covariance matrix $\mathbf{\Sigma}$ and the multivariate Gaussian distribution
$\mathbf{X}(\omega) \sim \boldsymbol{\mathcal{N}}(\boldsymbol{\mu},
\mathbf{\Sigma})$, defined by its multivariate CDF. The uncertainty in the
microstructure $\mathbb{M}(\omega)$ described in \autoref{eq:microstructure} is
defined by its uncertain fiber volume fraction $c_f(\omega)$ as seen in
\autoref{eq:microstructure_uniform} and in this work is considered  normally
distributed for UQ, i.e. $c_f(\omega) \sim \mathcal{N}$.

\ah{\textit{Remark 3:} In practice, the tails of Gaussian distributions lead to
    physically inadmissible values for the input values in \autoref{algo:uq}, as
    it has support on the entire real line $\mathbb{R}$. To circumvent this
    problem, in this work truncated Gaussian distributions
    \cite{burkardt2014truncated} have been used, which have bounded support.
    The tails are bounded, where physically meaningfull, i.e. between $0$ and
    $1$ for the volume fraction $c_f(\omega)$ and between $0$ and $0.5$ for the
    Poisson ratio, for lower and upper bounds, respectively.  For the shear
modulus $G$ and bulk modulus $K$, only strictly positive values are permitted.
Furthermore is is pointed out, that the random variables $\mathbf{X}(\omega)$ 
used in \autoref{algo:uq} have no connection to the random variables 
$\mathbf{X}^U(\omega)$ used for sampling the training data in
\autoref{algo:data_creation}.}

~\\~\\ \textit{(i) Cubature rule} \label{sec:Cubature_Rule} ~\\ 
Since Gaussian
random variables are used as input to Algorithm \ref{algo:uq}, 
\ah{Gauss-Hermite}
cubature is chosen \cite{abramowitz1988handbook} for numerical integration of
\autoref{eq:main_problem2} in Algorithm \ref{algo:uq}.i. \textit{ChaosPy}
\cite{feinberg_chaospy:_2015} uses Stieltjes method on three-term recurrence
coefficients for sampling. The order $n_w - 1$ of the cubature in
\autoref{eq:main_problem2} must be provided. Nodes ${\Theta}^{(j)}$ and
corresponding weights $w(\Theta^{(j)})$ are than chosen automatically with
respect to the number of univariate random variables, resulting in a total of
\ah{$n_q$} nodes as defined in \autoref{eq:dataset_cub}.

~\\ \textit{(ii) Deterministic solutions} \label{sec:Deterministic_Solutions}
~\\ The deterministic solutions for \autoref{eq:main_problem2} are carried out
on the nodes $\Theta^{(j)}$ by the trained ANN $\tilde{\mathcal{M}}_{ANN}$ in
Algorithm \ref{algo:uq}.ii. For given microstructure $\mathbb{M}(\omega)$
defined in \autoref{eq:microstructure}, six strain states are evaluated by the
trained ANN $\tilde{\mathcal{M}}_{{ANN}}$, as described in
\autoref{eq:homogenization}. For uncertain fiber volume fraction $c_f(\omega)$
in \autoref{eq:microstructure_uniform}, for every node $\Theta^{(j)}$ in
\autoref{eq:dataset_cub} a corresponding microstructure is generated
synthetically via \autoref{sec:Data_Creation}.

~\\~\\ \textit{(iii) Orthonormal polynomials}
\label{sec:Orthonormal_Polynomials} ~\\ For Gaussian distributions, as in the
case of the present work for the input of Algorithm \ref{algo:uq}, Hermite
polynomials $\Psi_{\mathbf{i}}$ are used in \autoref{eq:main_problem}. The
polynomial degree $n_{PCE}$ in \autoref{eq:surrogate_pce} must be provided.
Then, Polynomials are generated by three-term recurrence described in
\cite{xiu2010numerical}.

~\\~\\ \textit{(iv) PC coefficients} \label{sec:PC_Coefficients} ~\\ The PC
coefficients $\rv{\hat{\mathbb{C}}_{\mathbf{i}}^{ANN}}$ 
in \autoref{eq:main_problem2} are
then calculated by Algorithm (\ref{algo:uq}.iv). With the calculated PC
coefficients $\rv{\hat{\mathbb{C}}_{\mathbf{i}}^{ANN}}$ 
and the orthonormal polynomials
$\Psi_{\mathbf{i}}$, the uncertain effective macro elasticity tensor
$\rv{\bar{\mathbb{C}}_{ANN}(\omega)}$ 
is fully defined by \autoref{eq:main_problem2}.

~\\~\\ \textit{(v) Statistics} \label{sec:Statistics} ~\\ The central moments
like mean $\boldsymbol{\mu}_{\mathbf{Y}}$ in \autoref{eq:moment1} and standard
deviation $\boldsymbol{\sigma}_{\mathbf{Y}}$ in \autoref{eq:moment2} of
$\rv{\bar{\mathbb{C}}_{ANN}(\omega)}$ in 
\autoref{eq:main_problem2} can then be obtained
from the PC coefficients $\rv{\hat{\mathbb{C}}_{\mathbf{i}}^{ANN}}$
\autoref{eq:main_problem2}, which is carried out by \textit{ChaosPy}
\cite{feinberg_chaospy:_2015}. Corresponding PDF and CDF can be reconstructed by
kernel density estimators \cite{feinberg_chaospy:_2015}.

\section{Numerical examples} \label{sec:Examples} \subsection{Example 1:
effective transversely isotropic properties of carbon fiber reinforced polymer}
\label{sec:example_1} \subsubsection{Problem description} 
\begin{figure}[htb] \centering
    \subfloat[\label{fig:single_rve0}]{\includegraphics[width=0.45\textwidth]
    {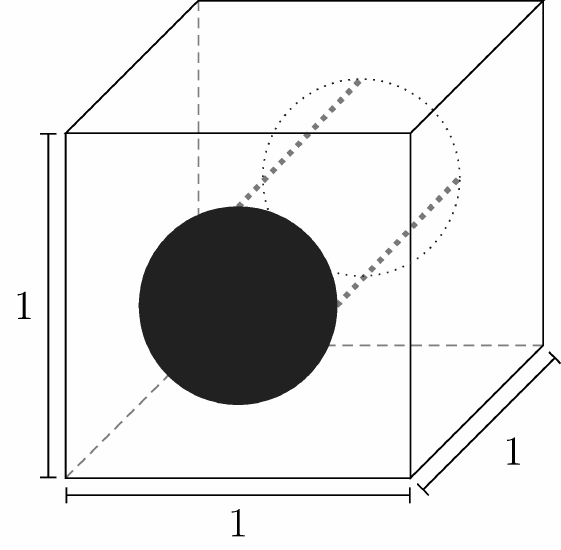}}
    \hfill
    \subfloat[\label{fig:single_rve}]{\includegraphics[width=0.45\textwidth]
    {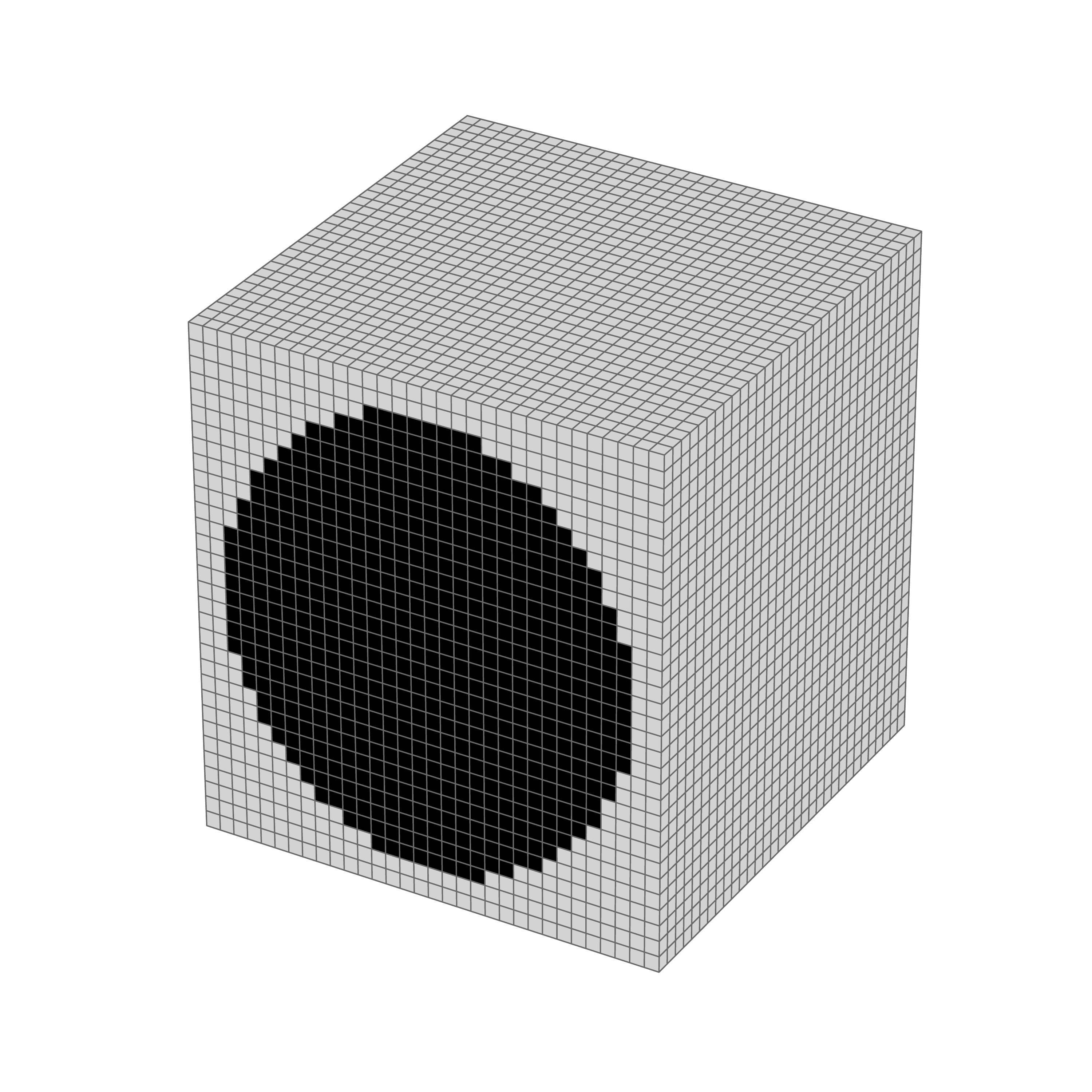}}
    \caption{Example 1: (a) Geometry of a long fiber inclusion centered in a
    matrix material. (b) Voxel discretization using 32 voxels per dimension.}
\end{figure}

The first example
compares the proposed method in \autoref{sec:Algorithmic_Procedure} with a
different uncertain homogenization approach from the literature. In
\cite{caylak_polymorphic_nodate}, the authors investigate a single long fiber
inclusion centered in a matrix material, where the cubic representative volume
element with unit dimensions $\xi_i$ in \autoref{eq:microstructure} is shown in
Figure \autoref{fig:single_rve0}. Here, deterministic fiber material parameters
$\boldsymbol{\kappa}^{I}$ and uncertain, normally distributed matrix material
parameters $\boldsymbol{\kappa}^{M}(\omega) \sim
\boldsymbol{\mathcal{N}}(\boldsymbol{\mu}, \boldsymbol{\sigma}^2)$ in
\autoref{eq:matpar} are considered. Additionally, uncertainty in the geometry is
studied by employing an uncertain, normally distributed fiber volume fraction
$c_f(\omega) \sim \mathcal{N}(\boldsymbol{\mu}, \boldsymbol{\sigma}^2)$ of the
microstructure $\mathbb{M}(\omega)$ in \autoref{eq:microstructure}, effectively
altering the radius of the fiber inclusion. The normal distributions
$\boldsymbol{\mathcal{N}}(\boldsymbol{\mu}, \boldsymbol{\sigma}^2)$ are fully
defined by their mean values $\boldsymbol{\mu}_{\mathbf{X}}$ in
\autoref{eq:mean} and standard deviations $\boldsymbol{\sigma}_{\mathbf{X}}$ in
\autoref{eq:deviation}. Given uncertain linear elastic isotropic material
parameters of single constituents in Table \ref{tab:results_2}, uncertain
effective transversal isotropic properties $\bar{E}_1(\omega),
\bar{E}_2(\omega), \bar{G}_{12}(\omega),  \bar{G}_{23}(\omega),
\bar{\nu}_{12}(\omega), \bar{\nu}_{23}(\omega)$ calculated from
$\rv{\bar{\mathbb{C}}_{ANN}(\omega)}$ 
in \autoref{eq:main_problem2} are of interest
\begin{equation} \bar{\mathbb{C}}(\omega) \xrightarrow{\text{Voigt}}
    \underline{\underline{\bar{\mathbb{C}}}}(\omega) =
    \left[\begin{array}{cccccc} \bar{\mathbb{C}}_{1111} & 2
            \bar\nu_{12}\left(\bar\lambda+\bar{G}_{23}\right) & 2
            \bar\nu_{12}\left(\bar\lambda+\bar{G}_{23}\right) & 0 & 0 & 0 \\ &
            \bar\lambda+2 \bar{G}_{23} & \bar\lambda & 0 & 0 & 0 \\ & &
            \bar\lambda+2 \bar{G}_{23} & 0 & 0 & 0 \\ & & & \bar{G}_{23} & 0 & 0
            \\ & \text { sym. } & & & \bar{G}_{12} & 0 \\ & & & & & \bar{G}_{12}
\end{array}\right] 
\label{eq:effective}
\end{equation} \begin{equation*} \bar{\mathbb{C}}_{1111} =
    \frac{1-\bar\nu_{23}}{1-\bar\nu_{23}-2 \bar\nu_{12} \bar\nu_{21}}
    \bar{E}_{1},\qquad \bar\lambda = \frac{\bar\nu_{12}
\bar\nu_{21}+\bar\nu_{23}}{\left(1-\bar\nu_{23}-2 \bar\nu_{12}
\bar\nu_{21}\right)\left(1+\bar\nu_{23}\right)} \bar{E}_{2} 
\end{equation*} where the elementary event $\omega$ in \autoref{eq:input}
applies to all variables, but is omitted for readability. In
\cite{caylak_polymorphic_nodate}, a Galerkin PCE with FEM for uncertain
full-field homogenization of the representative volume element shown in Figure
\autoref{fig:single_rve0} is utilized. Periodic boundary conditions are used, as
explained in \autoref{eq:BC}. 

In the following, the results from \cite{caylak_polymorphic_nodate} are used as
comparison to the proposed approach in \autoref{sec:Algorithmic_Procedure}. The
input values in Table \ref{tab:results_2} are the same as in
\cite{caylak_polymorphic_nodate}. 

\begin{figure}[htb]
    \subfloat[\label{fig:single_rve01}]{\includegraphics[width=0.3\textwidth]
    {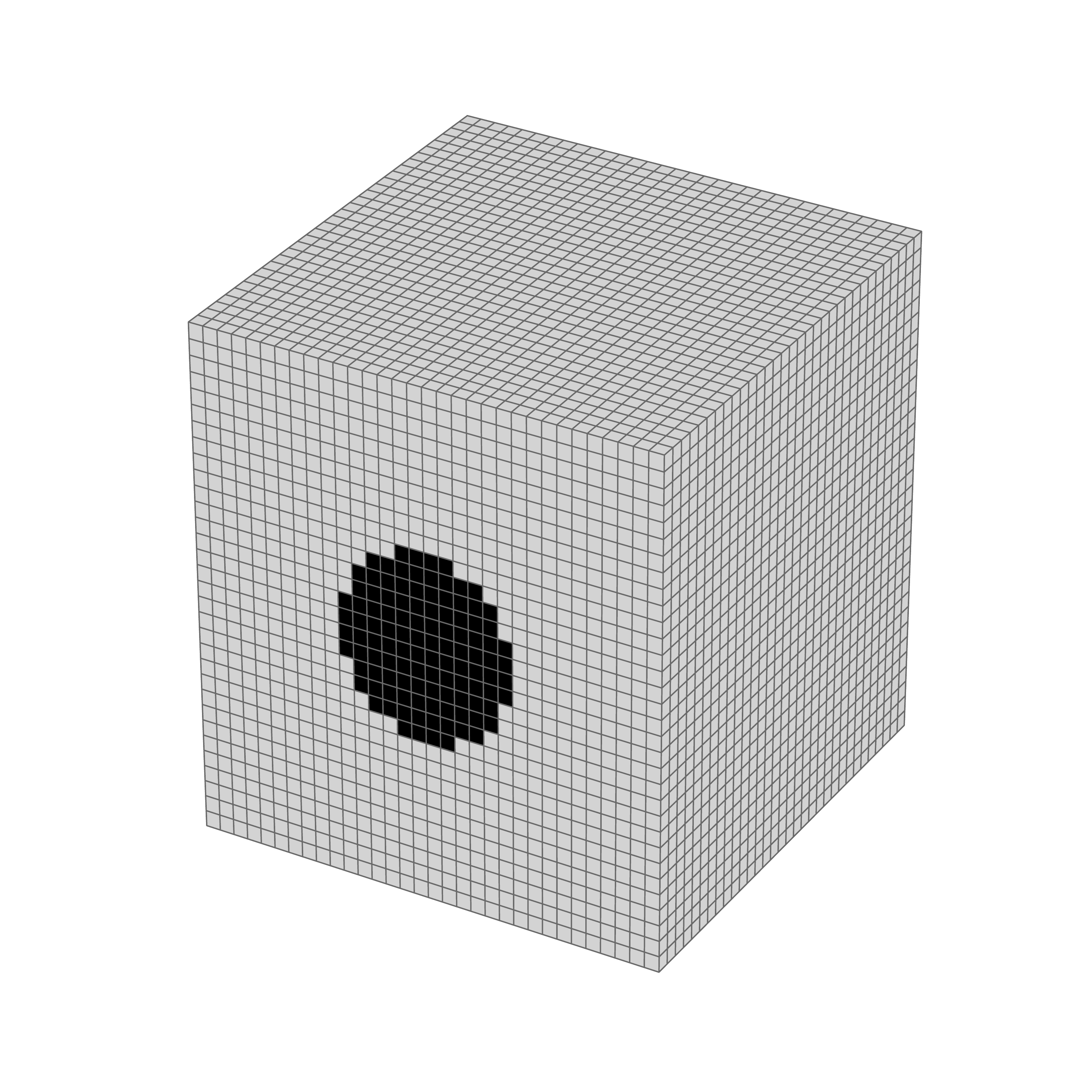}}
    \hfill
    \subfloat[\label{fig:single_rve04}]{\includegraphics[width=0.3\textwidth]
    {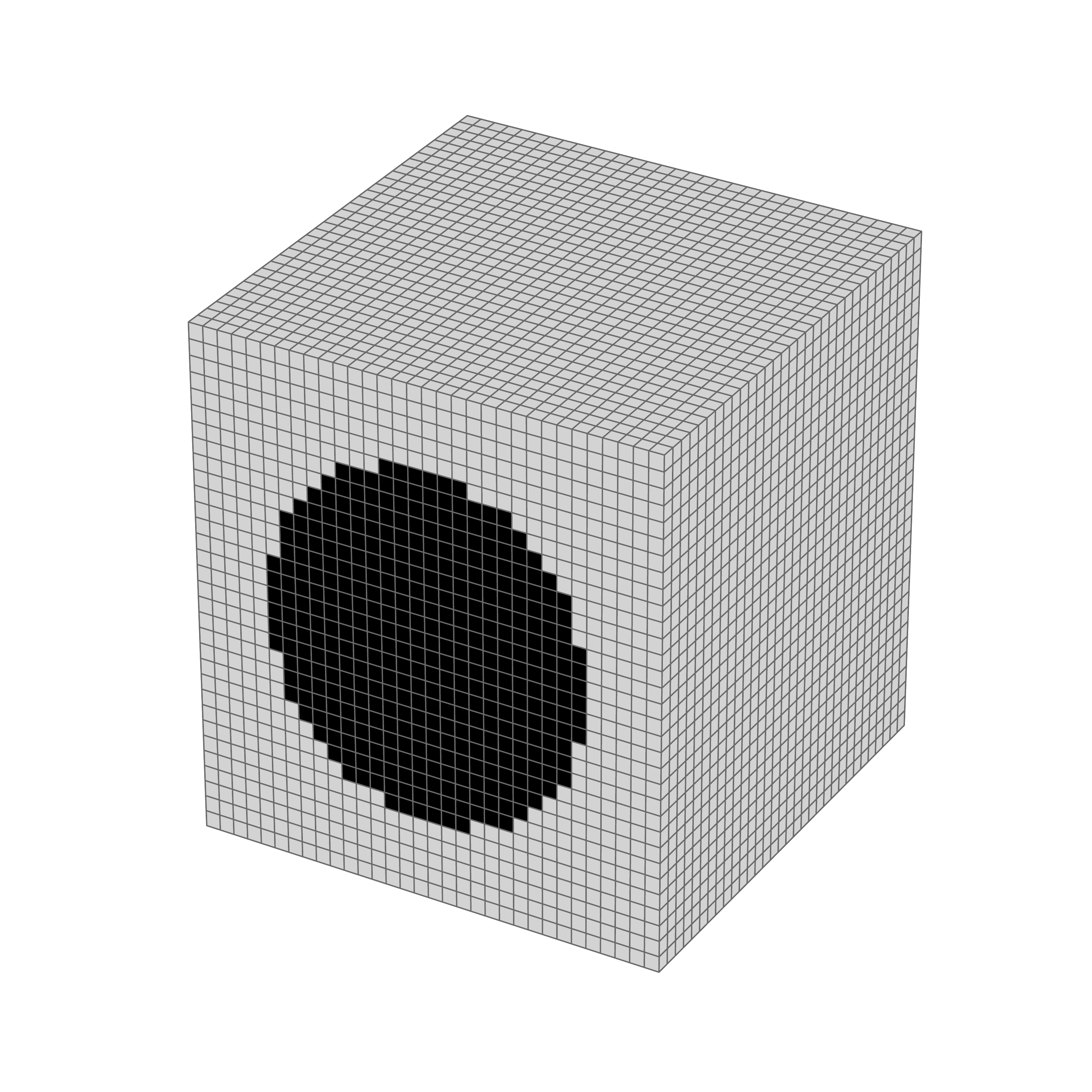}}
    \hfill
    \subfloat[\label{fig:single_rve07}]{\includegraphics[width=0.3\textwidth]
    {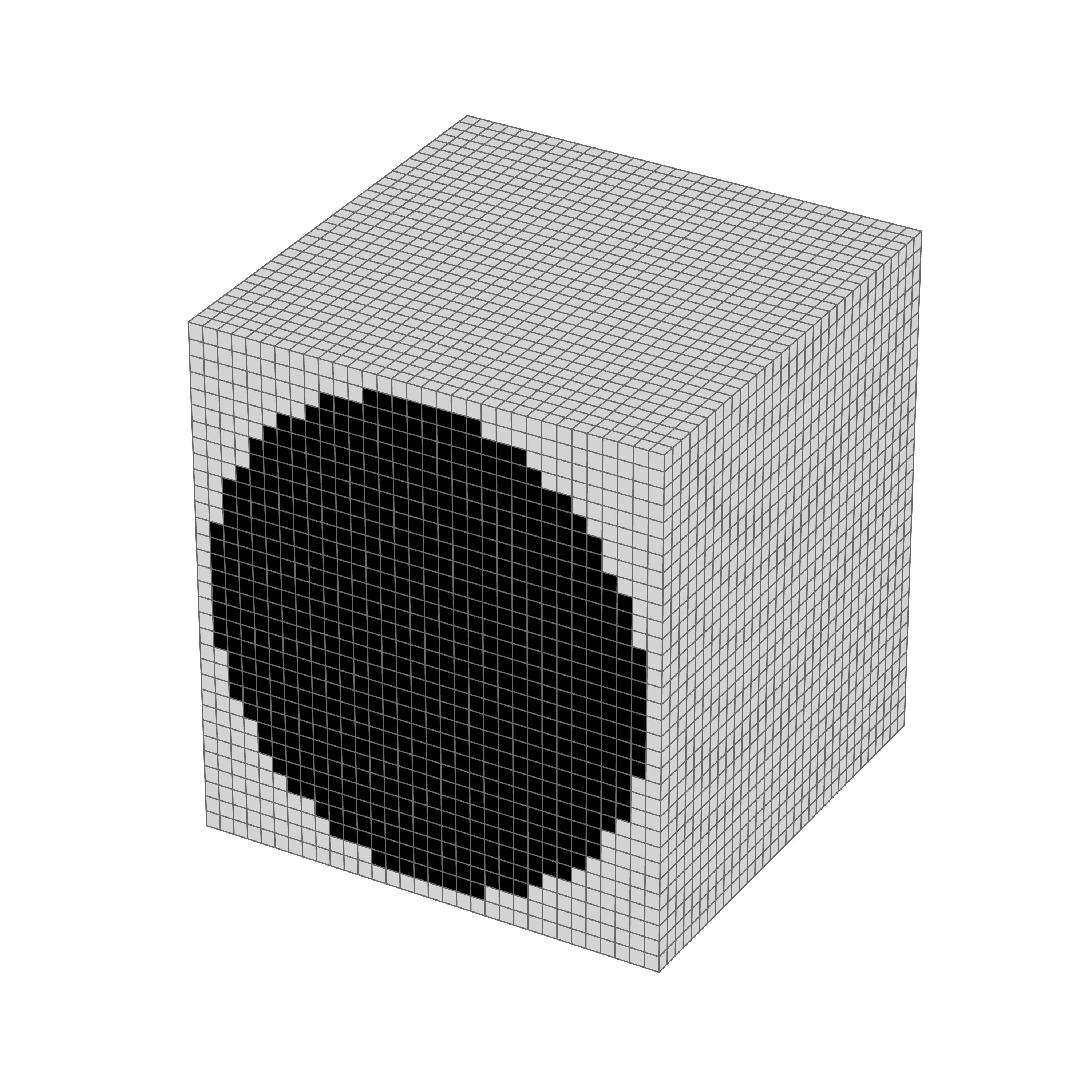}}
    \caption{Example 1: Geometry of a long fiber inclusion centered in a matrix
    material for different fiber volume fractions $c_f$ in
\autoref{eq:microstructure_uniform}. (a) $c_f = 0.1$, (b) $c_f = 0.4$, (c) $c_f
= 0.7$.} \end{figure}

\begin{table}[htb] \center \begin{tabular}{@{}clllll@{}} \toprule Parameters of
        $\mathcal{N}(\boldsymbol{\mu}_{\mathbf{X}},
        \boldsymbol{\sigma}_{\mathbf{X}}^2)$ & $c_f(\omega)$ [-] &
        $E^{M}(\omega) [\text{MPa}]$ & $\nu^{M}(\omega)$ [-] & $E^{I}
        [\text{MPa}]$ & $\nu^{I}$ [-]\\ \midrule $\boldsymbol{\mu}_{\mathbf{X}}$
        & $0.6335$ & $3.101 \times 10^3$ & $0.41$ & $2.31 \times 10^5$ & $0.1$\\
     $\boldsymbol{\sigma}_{\mathbf{X}}$ & $0.0264$ & $1.11 \times 10^2$ &
 $0.044$& - & - \\ \bottomrule \end{tabular} \caption{Example 1: Parameters for
 the normally distributed input random variable $\mathbf{X}(\omega) \sim
 \boldsymbol{\mathcal{N}}(\boldsymbol{\mu}_{\mathbf{X}},
 \boldsymbol{\sigma}_{\mathbf{X}}^2)$ of the dataset $\mathbb{D}_{ANN}$
 according to the input of Algorithm \ref{algo:uq} and \autoref{eq:moment1},
 \autoref{eq:moment2}.} \label{tab:results_2} \end{table}

\subsubsection{Data generation for deep learning using FFT} Following
\autoref{sec:Data_Creation} and Algorithm \ref{algo:data_creation}, a dataset
$\mathbb{D}_{ANN}$ as defined in \autoref{eq:dataset_ann} is created. Here, $n_s
= 9000$ three dimensional microstructures $\mathbb{M}(\omega)$ in
\autoref{eq:microstructure}, discretized by voxels as shown in Figure
\autoref{fig:single_rve01} - (c), are homogenized by FFT on a computer cluster.
The uniform distributions from \autoref{tab:results_1} are used for sampling the
\rv{inputs $c_{f}$ and $\boldsymbol{\kappa}^{M} = \{E^M,
\nu^M\}$, where the fiber parameters $\boldsymbol{\kappa}^{I}$ are
fixed. The upper and lower boundaries $a$ and $b$, respectively, are chosen with
physical constraints in mind as explained in \autoref{eq:phyconst}.}

~\\~\\ \ah{\textit{Remark 4:} If the fiber volume fraction $c_f$ 
        approaches 1,
the stiffness of the microstructure will be identical to the deterministic
stiffness of the fiber. For $c_f$ close to 1, overlapping of the fiber
can appear. This unphysical behavior can be avoided by carefully choosing the
fiber volume fraction in the stochastic analysis in Section 4.1.3
, which should be well below 1.}

\begin{table}[htb] \center \begin{tabular}{@{}clll@{}} \toprule Parameters of
        $\mathcal{U}(\underbar{x}, \bar{x})$ 
        & \rv{$c_f$ [-]}& \rv{$E^{M} [\text{MPa}]$} & \rv{
        $\nu^{M}$ [-]} \\ 
        \midrule 
        $\underbar{x}$ & 0 & $10^3$ & 0.1 \\ 
        $\bar{x}$ & 1 & $10^4$
        & 0.48 \\ \bottomrule \end{tabular} \caption{Example 1: Uniform
        distributions with lower and upper limit $(a, b)$, respectively, of
    training samples $\mathbf{x}_k$
according to \autoref{eq:input2}.} 
\label{tab:results_1} \end{table}

\subsubsection{ANN design and training} An ANN is created following Section
\ref{sec:ANN_Creation}, Algorithm 3 and according to the topology outlined in
\autoref{fig:ANN_topo}. For the CNN in \autoref{fig:ANN_topo}, an
\textit{AlexNet} is chosen, as described in Section \ref{sec:Topology}.(i). The
hyperparameters of the ANN according to \autoref{tab:hp} are chosen by random
search following Section \ref{sec:Hyperparameter}.(ii). The optimized values,
which give the lowest error on a corresponding dataset $\mathbb{D}_{HP}$, are
given in \autoref{tab:hp_1}. The mean relative error after training with respect
to a test set using these hyperparameters is $2.54\%$.
%

\begin{table}[htb] \center \begin{tabular}{@{}lllllll@{}} \toprule Symbol &
    $\alpha$ & $\beta$ & $\lambda_{L2}$ & $n_u$ & $n_F$ & $n_L$ \\ \midrule
Optimized value & 0.005 & 0.0 & 0.0 & 2048 & 32 & 2 \\ \bottomrule \end{tabular}
\caption{Example 1: Optimized values for hyperparameters of the ANN from
\autoref{tab:hp} with respect to the error function in
\autoref{eq:optimization}, which lead to the lowest error $\mathcal{E}$.}
\label{tab:hp_1} \end{table}
\subsubsection{UQ using PCE and ANN trained on FFT} After training of the ANN
$\tilde{\mathcal{M}}_{ANN}$ in the previous section, UQ is carried out following
Section \ref{sec:UQ} and Algorithm \ref{algo:uq}. The sample distributions of
the input parameters $\mathbf{X}(\omega)$ in Algorithm \ref{algo:uq} are
according to \cite{caylak_polymorphic_nodate}, which are outlined in
\autoref{tab:results_2}. Consequently, the dimension $n_x$ of the random input
space $\mathcal{D}_{\mathbf{X}}$ is $n_x = |\{c_f(\omega), E^M(\omega),
\nu^M(\omega)\}| = 3$. For integration a Gaussian multivariate cubature of order
$n_w - 1 = 9$ in \autoref{eq:pseudo} is chosen. The calculations of $\ah{n_q} =
(n_w=10)^{n_x=3} = 10^3$ deterministic solutions in \autoref{eq:main_problem2}
are carried out by the trained ANN $\rv{\tilde{\mathcal{M}}_{ANN}^t}$. Hermite
polynomials $\Psi_{\mathbf{i}}$ with order $n_{PCE} = 9$ are used as polynomial
basis in \autoref{eq:surrogate_pce}. The PC coefficients
$\hat{\mathbb{C}}_{\mathbf{i}}$ are then calculated by pseudospectral PCE
defined in \autoref{eq:surrogate_pce}. 

\begin{figure}[htb] \centering
    \subfloat{\includegraphics[width=\textwidth]{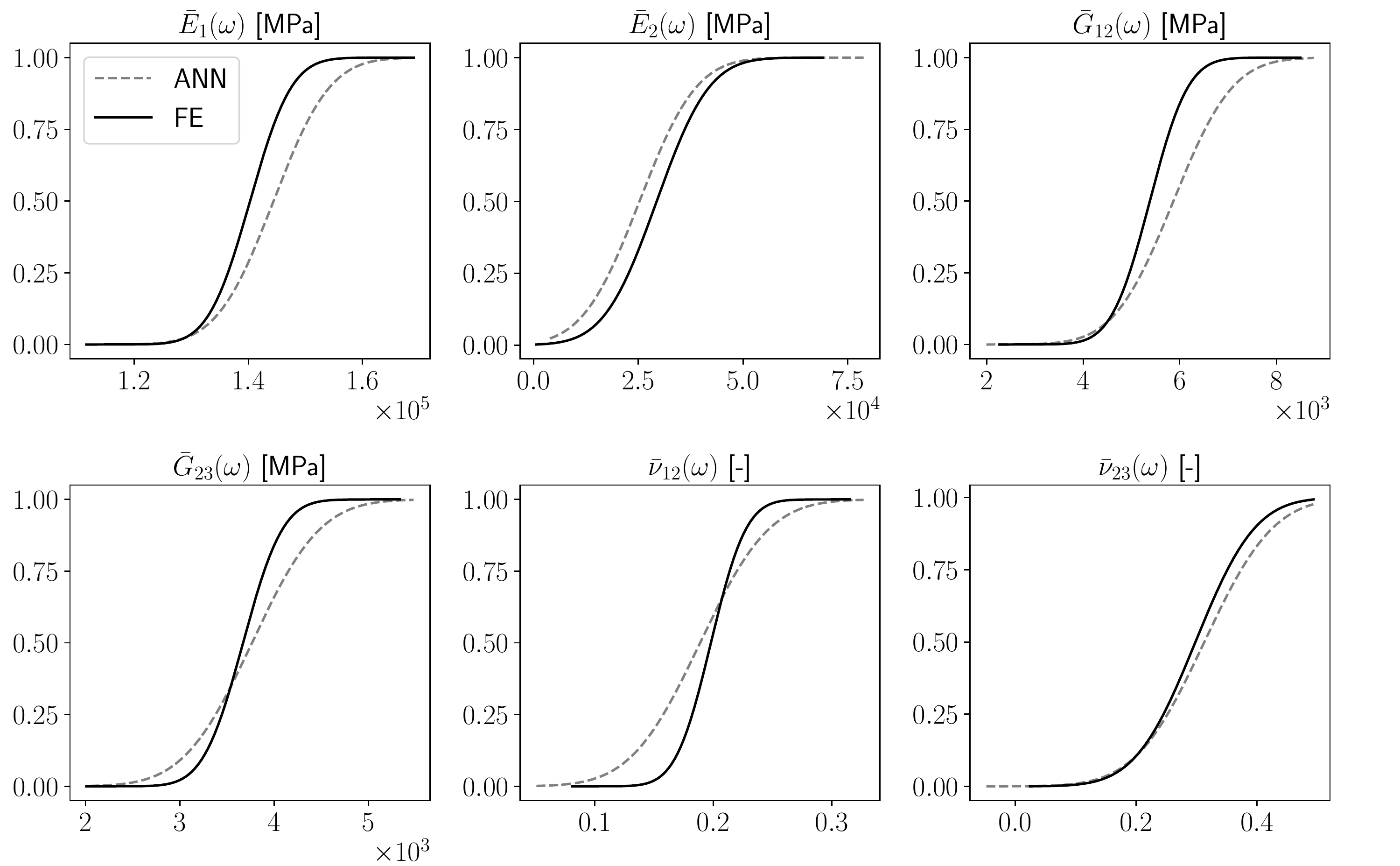}} \caption{Example
    1: Comparison of CDFs of uncertain effective transversal isotropic
properties $\bar{E}_1(\omega), \bar{E}_2(\omega), \bar{G}_{12}(\omega),
\bar{G}_{23}(\omega), \bar{\nu}_{12}(\omega), \bar{\nu}_{23}(\omega)$ in
\autoref{eq:effective} of the ANN model (ANN) and the finite element reference
solution by \cite{caylak_polymorphic_nodate} (FE). } \label{fig:caylak}
\end{figure}

The resulting CDFs defined in \autoref{eq:surrogate_pce} of the uncertain
effective transversal isotropic properties $\bar{E}_1(\omega),
\bar{E}_2(\omega), \bar{G}_{12}(\omega),  \bar{G}_{23}(\omega),
\bar{\nu}_{12}(\omega), \bar{\nu}_{23}(\omega)$ in \ah{ \autoref{eq:effective}} 
of the
uncertain effective elasticity tensor $\rv{\bar{\mathbb{C}}_{ANN}(\omega)}$ in
\autoref{eq:main_problem2} are shown in \autoref{fig:caylak}. The solutions of
the proposed algorithm are denoted by \textit{ANN} whereas the reference
solutions obtained by \cite{caylak_polymorphic_nodate} using FEM are denoted by
\textit{FE}. Between these two methods, an agreement of the corresponding CDFs
can be observed. \ah{It has to be mentioned, that deviations in CDFs can be more
    pronounced in PDFs. For consistency with \cite{caylak_polymorphic_nodate} as
    well as unified presentation, 
in this work CDFs were chosen to display results.} 
Deviations can be explained by multiple factors. First, the
voxel discretization shown in \autoref{fig:single_rve} differs from the finite
element discretization in \cite{caylak_polymorphic_nodate}. As can be seen in
Figure \autoref{fig:single_rve01} - c, circular geometries are only approximated
voxels, resulting in stair-like effects. These differences lead to perturbations
in the micro stress field $\boldsymbol{\sigma}(\omega)$ in
\autoref{eq:uequilibrium}, which ultimately influence the effective macro
properties of the uncertain effective elasticity tensor
$\bar{\mathbb{C}}(\omega)$ in \autoref{eq:main_problem2} using
\autoref{eq:homogenization} utilizing FFT in \autoref{eq:fft}. Second, the
approximation error of the ANN in \autoref{eq:optimization} leads to errors in
the deterministic solution $\rv{\tilde{\mathcal{M}}_{\text{ANN}}^t}$ of
\autoref{eq:main_problem2}. Third, the pseudospectral approach of PCE in
\autoref{eq:pseudo} inherits a number of approximations contributing to the
overall deviations between both homogenization approaches, namely usage of
cubature and different polynomial orders of the orthonormal polynomials in
\autoref{eq:surrogate_pce}.

Despite the minor deviations in the CDFs, the proposed algorithm is capable of
predicting the uncertain effective properties of transversely isotropic fiber
reinforced materials.

 \subsection{Example 2: effective isotropic properties of spherical inclusions}
\label{sec:example_2} \subsubsection{Problem description} 
\begin{figure}[htb] \centering
    \subfloat[\label{fig:sphere_rve0}]{\includegraphics[width=0.45\textwidth]
    {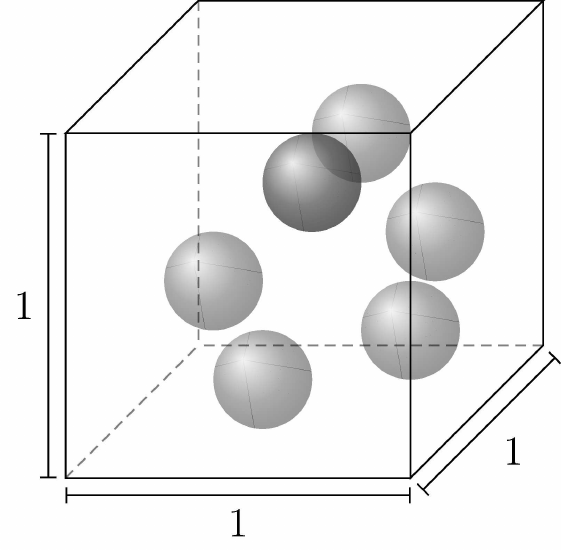}}
    \hfill
    \subfloat[\label{fig:sphere_rve}]{\includegraphics[width=0.45\textwidth]
    {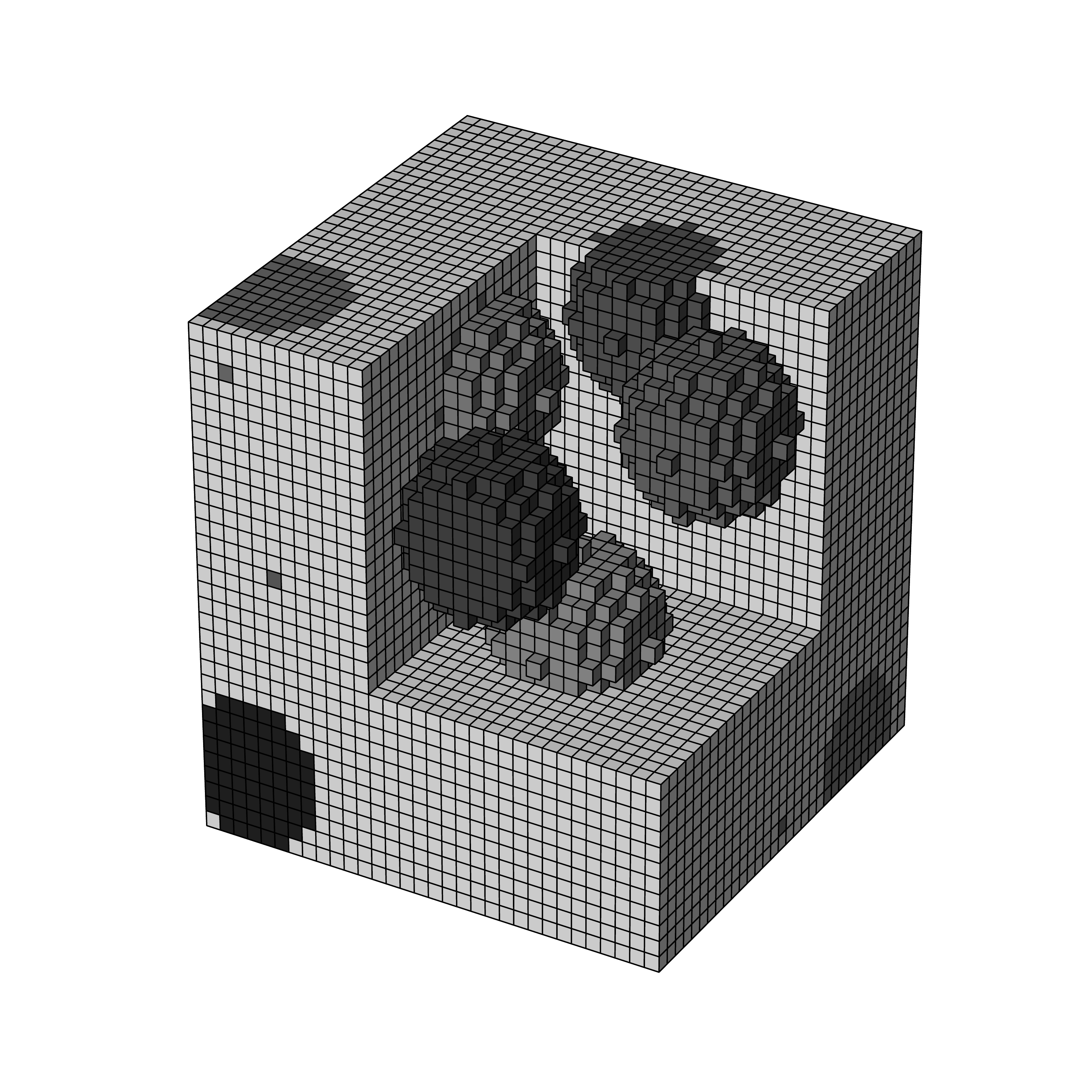}}
    \caption{Example 1: (a) Geometry of multiple spherical inclusion embedded in
    a matrix material. (b) Voxel discretization using 32 voxels per dimension.}
\end{figure}

The second example
deals with more complex microstructure compared to Example 1. A cubical unit
cell consisting of matrix material with multiple spherical inclusions, as shown
in Figure \autoref{fig:sphere_rve}, is considered. The performance of the
proposed method is compared to MC simulations of a deterministic FFT solver, in
particular the speed up of the ANN approach is investigated. \ah{Additionally, 
    to
compare the error induced by the PCE, MC simulations using the ANN solver are
carried out.} In this example,
the Matrix material parameters $\boldsymbol{\kappa}^{M}(\omega)$ in
\autoref{eq:matpar} are normally distributed random variables. The corresponding
parameters are shown in Table \ref{tab:2}. The parameters of the inclusion
material $\boldsymbol{\kappa}^{I}$ are considered to be deterministic.
Uncertainty of the geometry is taken into account by a normally distributed 
\ah{inclusion} volume fraction $c_f(\omega) \sim \mathcal{N}(\boldsymbol{\mu},
\boldsymbol{\sigma}^2)$ of the microstructure $\mathbb{M}(\omega)$ in
\autoref{eq:microstructure}. This \ah{inclusion} volume fraction then 
determines the
number of spherical inclusions. Normal distributions
$\mathcal{N}(\boldsymbol{\mu}_{\mathbf{X}}, \boldsymbol{\sigma}_{\mathbf{X}}^2)$
again are completely defined by \autoref{eq:mean} and \autoref{eq:deviation}.
Because of the random placement of the spherical inclusions, the macroscopic
material behaviour is assumed to be isotropic. Therefore,  uncertain effective
linear elastic isotropic properties $\bar{E}(\omega), \bar{\nu}(\omega)$
calculated from $\rv{\bar{\mathbb{C}}_{ANN}(\omega)}$ 
in \autoref{eq:main_problem2} are of
interest

        \begin{equation} \bar{\mathbb{C}}(\omega) \xrightarrow{\text{Voigt}}
            \underline{\underline{\bar{\mathbb{C}}}}(\omega)
            =\left[\begin{array}{cccccc} 2 \bar\mu+\bar\lambda & \bar\lambda &
                    \bar\lambda & 0 & 0 & 0 \\ & 2 \bar\mu+\bar\lambda &
                    \bar\lambda & 0 & 0 & 0 \\ & & 2 \bar\mu+\bar\lambda & 0 & 0
        & 0 \\ & & & \bar\mu & 0 & 0 \\ & \text{sym.} & & & \bar\mu & 0 \\ & & &
        & & \bar\mu \end{array}\right] \end{equation} \begin{equation*}
            \bar{\mu} = \frac{\bar{E}}{2(1+\bar\nu)},\qquad \bar\lambda =
        \frac{\bar{E} \bar\nu}{(1+\bar\nu)(1-2 \bar\nu)} \label{eq:effective2}
    \end{equation*} where the elementary event $\omega$ \autoref{eq:input}
    applies to all variables but is omitted for readability. 
    \ah{Deviations from the isotropic form in \autoref{eq:effective2} are
    neglected.}
    For the reference
    MC simulations using the FFT solver, the number of samples is chosen as 
    $n_s = 10^3$. This number
    of simulations is sufficient for mean and standard deviation estimation as
    mentioned by \cite{sudret_surrogate_2017}. 
    \ah{For the MC simulations using the
    ANN solver, the number of samples is chosen as 
$n_s = 10^4$.} 

\ah{\textit{Remark 5:}
        The ANN has to be trained only once. After that, a deterministic
        surrogate for the FFT solver has been established, which can be used for
        multiple stochastic investigations. The computational effort
        is one time only and is front loaded. As the sample generation is
        pleasingly parallel, it is well suited for computation on cluster 
    computers. After training, the ANN is much faster than the original FFT
solver, as can be seen \autoref{fig:time}.}

\begin{figure}[htb]
    \subfloat[\label{fig:sphere_rve_005}]{\includegraphics[width=0.3\textwidth]
    {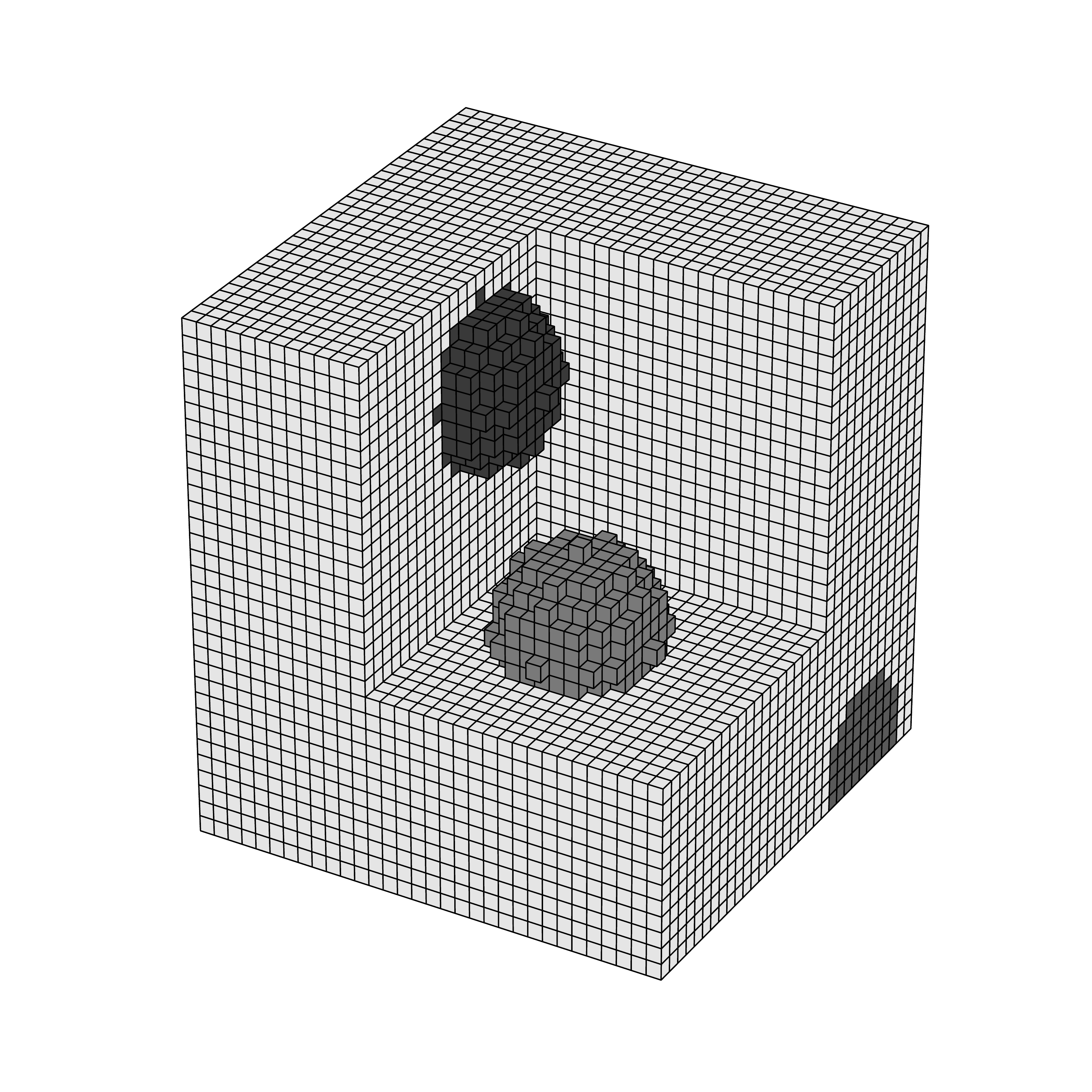}}
    \hfill
    \subfloat[\label{fig:sphere_rve_03}]{\includegraphics[width=0.3\textwidth]
    {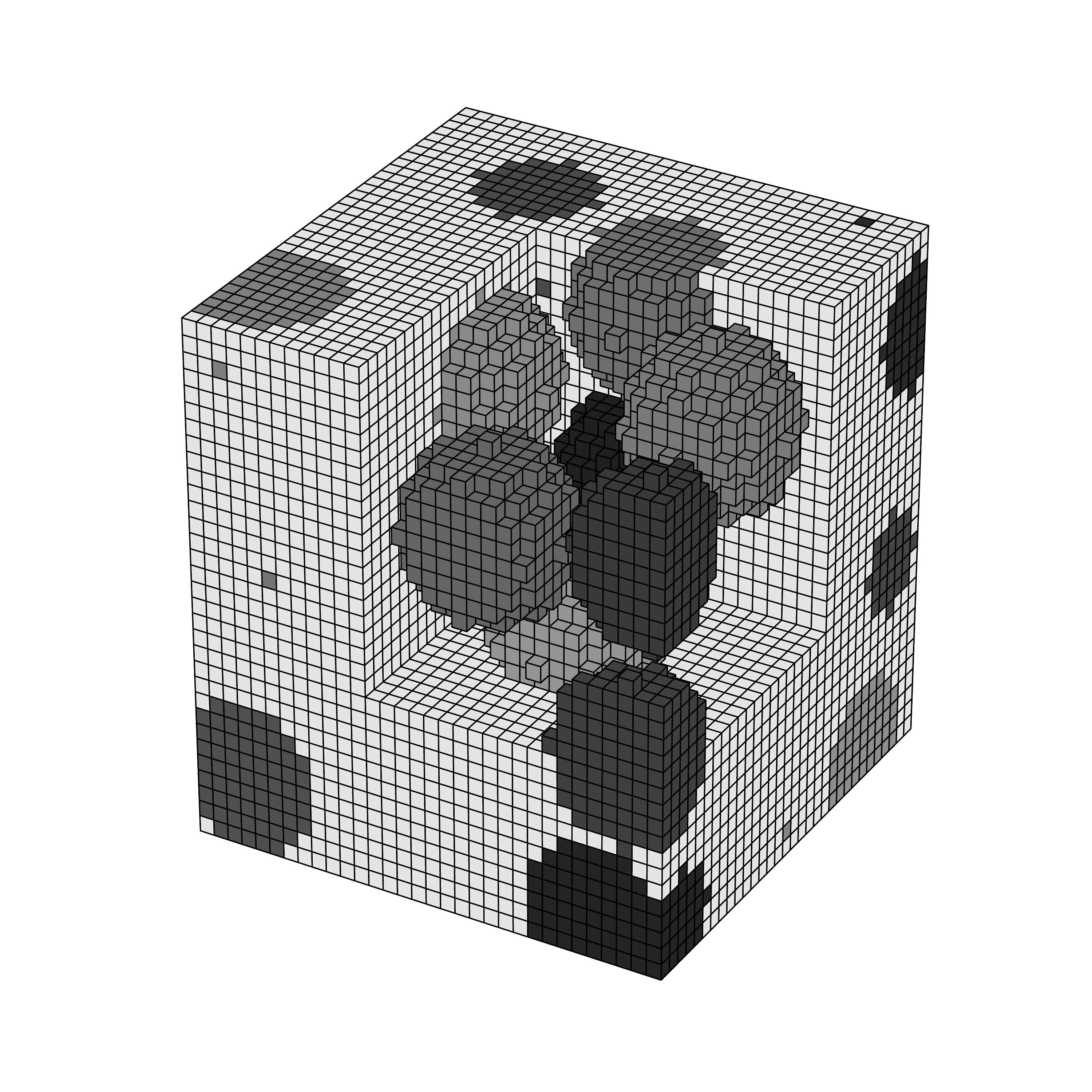}}
    \hfill
    \subfloat[\label{fig:sphere_rve_04}]{\includegraphics[width=0.3\textwidth]
    {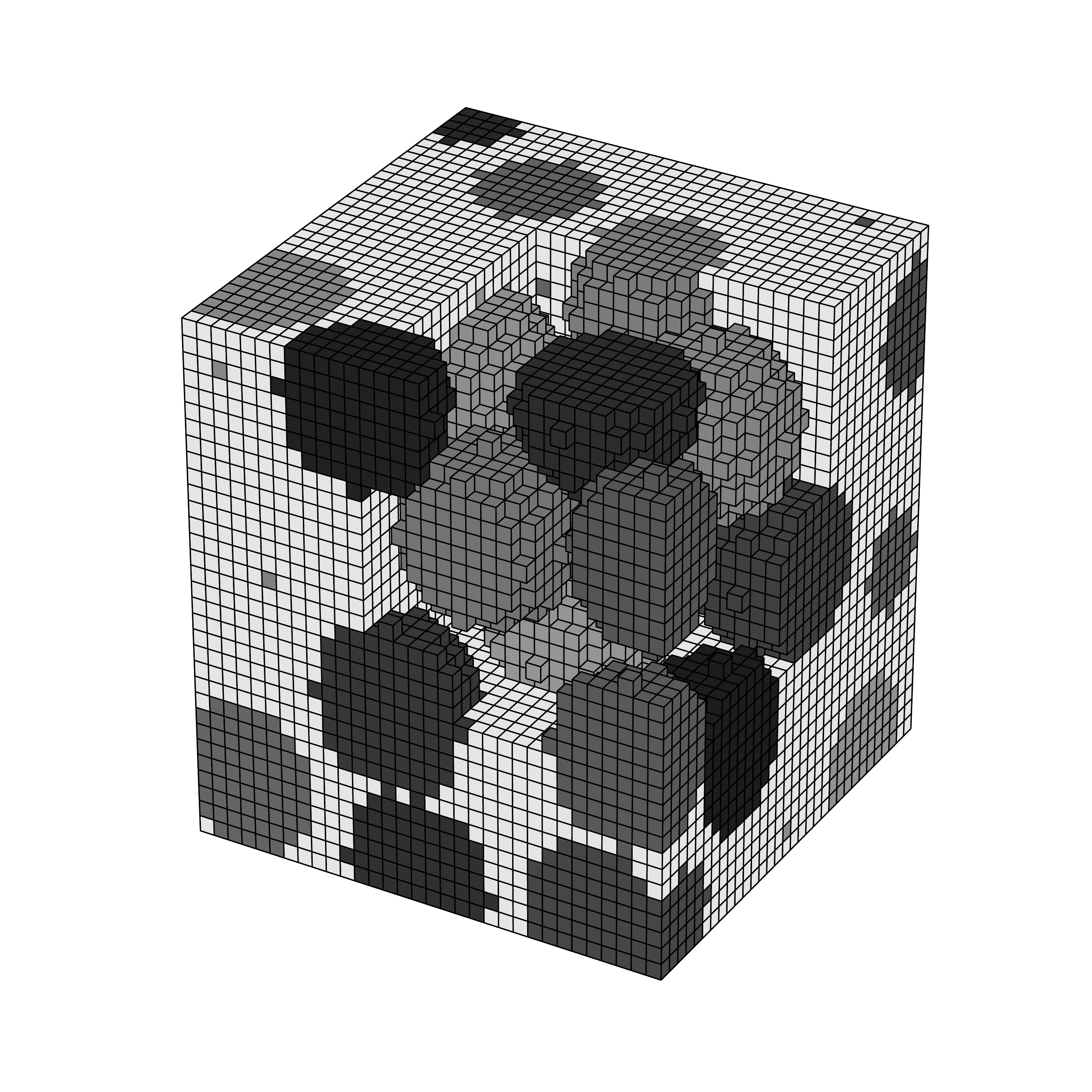}}
    \caption{Example 1: Geometry of multiple spherical inclusion embedded in a
        matrix material for different \ah{inclusion} volume fractions $c_f$ in
\autoref{eq:microstructure_uniform}. (a) $c_f = 0.05$. (b) $c_f = 0.3$. (c) $c_f
= 0.4$.} \end{figure}

\begin{table}[htb] \center \begin{tabular}{@{}clllll@{}} \toprule Parameters of
        $\mathcal{N}(\boldsymbol{\mu}_{\mathbf{X}},
        \boldsymbol{\sigma}_{\mathbf{X}}^2)$ & $c_f(\omega)$ [-] &
        $E^{M}(\omega) [\text{MPa}]$ & $\nu^{M}(\omega)$ [-] & $E^{I}
        [\text{MPa}]$ & $\nu^{I}$ [-]\\ \midrule $\boldsymbol{\mu}_{\mathbf{X}}$
        & $0.2$ & $ 5 \times 10^3$ & $0.3$ & $ 2.31 \times 10^5$ & $0.1$\\
        $\boldsymbol{\sigma}_{\mathbf{X}}$ & $0.02$ & $ 5 \times 10^2$ & $0.03$
        & - & - \\ \bottomrule \end{tabular} \caption{Example 2: Parameters for
        the normally distributed input random variables $\mathbf{X}(\omega) \sim
    \mathcal{N}(\boldsymbol{\mu}_{\mathbf{X}},
\boldsymbol{\sigma}_{\mathbf{X}}^2)$ of the dataset $\mathbb{D}_{ANN}$ according
to the input of Algorithm \ref{algo:uq} and \autoref{eq:moment1},
\autoref{eq:moment2}.} \label{tab:2} \end{table}

\subsubsection{Data generation for deep learning using FFT} Following
\autoref{sec:Data_Creation} and Algorithm \ref{algo:data_creation}, a dataset
$\mathbb{D}_{ANN}$ defined in \autoref{eq:dataset_ann} with $n_s = 13800$ three
dimensional microstructures $\mathbb{M}$ from
\autoref{eq:microstructure}, discretized by voxels as shown in Figure
\autoref{fig:sphere_rve_005} - c, is homogenized by FFT as defined in
\autoref{eq:fft} on a computer cluster. The uniform distributions from
\autoref{tab:results_2_uni} are used for sampling the \rv{inputs $c_f,
E^M$ and $\nu^M$, where the \ah{inclusion} parameters $\boldsymbol{\kappa}^{I}$
are
fixed. The upper and lower boundaries $a$ and $b$, respectively, are chosen with
physical constraints in mind as explained in \autoref{eq:phyconst}.} 
\ah{During microstructure generation, overlap of particles is
    avoided by the algorithm by rejecting spherical inclusion, where shared 
    voxels with already placed particles are detected.}
\subsubsection{ANN design and training} An ANN is created
following Section \ref{sec:ANN_Creation} and Algorithm 3, where the topology is
shown in \autoref{fig:ANN_topo}. In contrast to Example 1, for the CNN in
\autoref{fig:ANN_topo}, a 40-layer \textit{DenseNet} is chosen to account for
the more complex microstructure as described in Section \ref{sec:Topology}. The
hyperparameters of ANN according to \autoref{tab:hp} are chosen by random
search, similarly as for Example 1. The optimal values for the topology chosen
are given in \autoref{tab:hp_2}. The mean-relative error with respect to a test
set is 4.0\%. This higher error is explained by the more complex microstructure
as well as the choice of CNN, which is deeper and therefore more difficult to
train. Nevertheless, the \textit{DenseNet} performed better than
\textit{AlexNet} for the microstructure considered, although this comparison is
not shown in the present work.  

\begin{table}[htb] 
    \center 
    \begin{tabular}{@{}clll@{}} 
        \toprule 
        Parameters of $\mathcal{U}(\underbar{x}, \bar{x})$ & \rv{$c_f$ [-]}& 
        \rv{$E^{M} [\text{MPa}]$} &
    \rv{$\nu^{M}(\omega)$ [-]} \\ 
    \midrule $\underbar{x}$ & 0 & $10^3$ & 0.1 \\ 
    $\bar{x}$ & 0.4 &
        $10^4$ & 0.48 \\ \bottomrule \end{tabular} \caption{Example 2: Uniform
        distributions with lower and upper limit $a$ and $b$, respectively, of
    training samples $\mathbf{x}_k$ according to \autoref{eq:input2}.} 
\label{tab:results_2_uni} \end{table}

\begin{table}[htb] \center \begin{tabular}{lcccccc} \toprule Symbol & $\alpha$ &
    $\beta$ & $\lambda_{L2}$ & $n_u$ & $n_F$ & $n_L$ \\ \midrule Optimum & 0.01
& 0.15 & $10^{-4}$ & 1024 & - & 2 \\ \bottomrule \end{tabular} \caption{Example
2: Optimized values for hyperparameters of the ANN from \autoref{tab:hp} with
respect to the error function in \autoref{eq:optimization}, which lead to the
lowest error $\mathcal{E}$.} \label{tab:hp_2} \end{table}

\subsubsection{UQ using PCE and ANN trained on FFT} 
\begin{figure}[htb] \centering \subfloat{\includegraphics[width=\textwidth,
    height=0.5\textwidth]{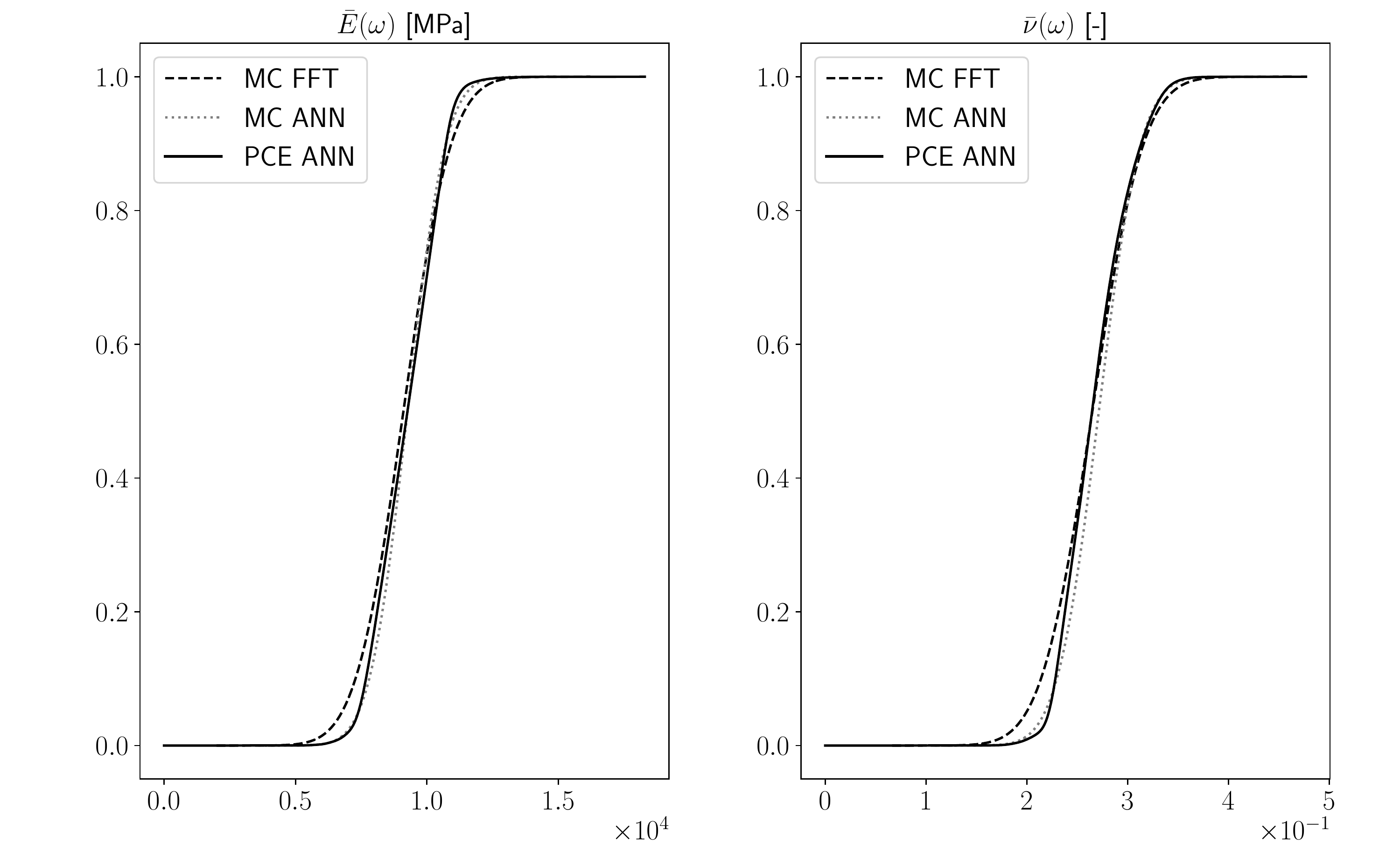}} 
    \caption{\ah{Example 2:
    Comparison of CDFs of uncertain effective linear elastic isotropic 
    properties $\bar{E}(\omega), \bar{\nu}(\omega)$ of the ANN model using 
    PCE (PCE ANN), the ANN model using $10^4$ MC simulations and $10^3$ MC
simulations of the FFT solver (MC FFT).}}
\label{fig:MC_1000} \end{figure}

The UQ is carried out
similarly to the procedure in Example 1. Following \autoref{sec:UQ} and
Algorithm \ref{algo:uq}, first the sample distributions of the multivariate
random input variables ${\mathbf{x}}_k = \left\{ \boldsymbol{\mathbb{M}},
\boldsymbol{\kappa},\bar{\underline{\boldsymbol{\varepsilon}}} \right\}$ in the
input of Algorithm \ref{algo:uq} are chosen according to \autoref{tab:2}. A
Gaussian multivariate cubature of order $n_w - 1 = 9$ in \autoref{eq:pseudo} is
chosen. Deterministic solutions are provided by the trained ANN
$\rv{\tilde{\mathcal{M}}_{ANN}^t}$ in \autoref{eq:main_problem2}. Hermite
polynomials with order $n_{PCE} = 9$ are used as defined in
\autoref{eq:surrogate_pce}. The
PC coefficients $\rv{\hat{\mathbb{C}}^{ANN}_{\mathbf{i}}}$ 
are calculated by pseudospectral
PCE defined in \autoref{eq:surrogate_pce}. The CDFs of the uncertain effective
linear elastic isotropic properties $\bar{E}(\omega), \bar{\nu}(\omega)$ from
\autoref{eq:effective2} of the uncertain effective elasticity tensor
$\rv{\bar{\mathbb{C}}_{ANN}(\omega)}$ 
from \autoref{eq:main_problem2} are shown in
\autoref{fig:MC_1000}. \ah{The solution from the ANN using PCE is denoted by 
\textit{PCE ANN}
while the reference solution from MC simulations using the FFT solver is 
denoted by \textit{MC FFT}. The solution from MC simulations using the ANN 
solver is denoted by \textit{MC ANN}.}

The comparison of the CDFs between the proposed method and the reference is
shown in \autoref{fig:MC_1000}. The results indicate, that the presented method
is capable of predicting uncertain effective properties for complex
microstructures consisting of matrix material and spherical, randomly
distributed spherical inclusions. The deviations can be explained similarly to
Example 1. For Example 2, the approximation error of the network is larger than
in the case of single fiber inclusions. Again, the usage of cubature and
polynomial expansion in the PCE yield further error sources, \ah{but these are
minor, as comparisons of the ANN using PCE and the ANN using MC in 
\autoref{fig:MC_1000} indicate. The mean values for
both effective properties show close correspondence to the reference FFT 
solution,
while their respective variances differ.} The asymmetrical CDF with respect to
the Poisson's ratio $\bar{\nu}(\omega)$ in \autoref{fig:MC_1000} indicates, that
the network accuracy alters for certain values. 

\begin{figure}[htb] \centering \subfloat{\includegraphics[width=\textwidth,
    height=0.5\textwidth]{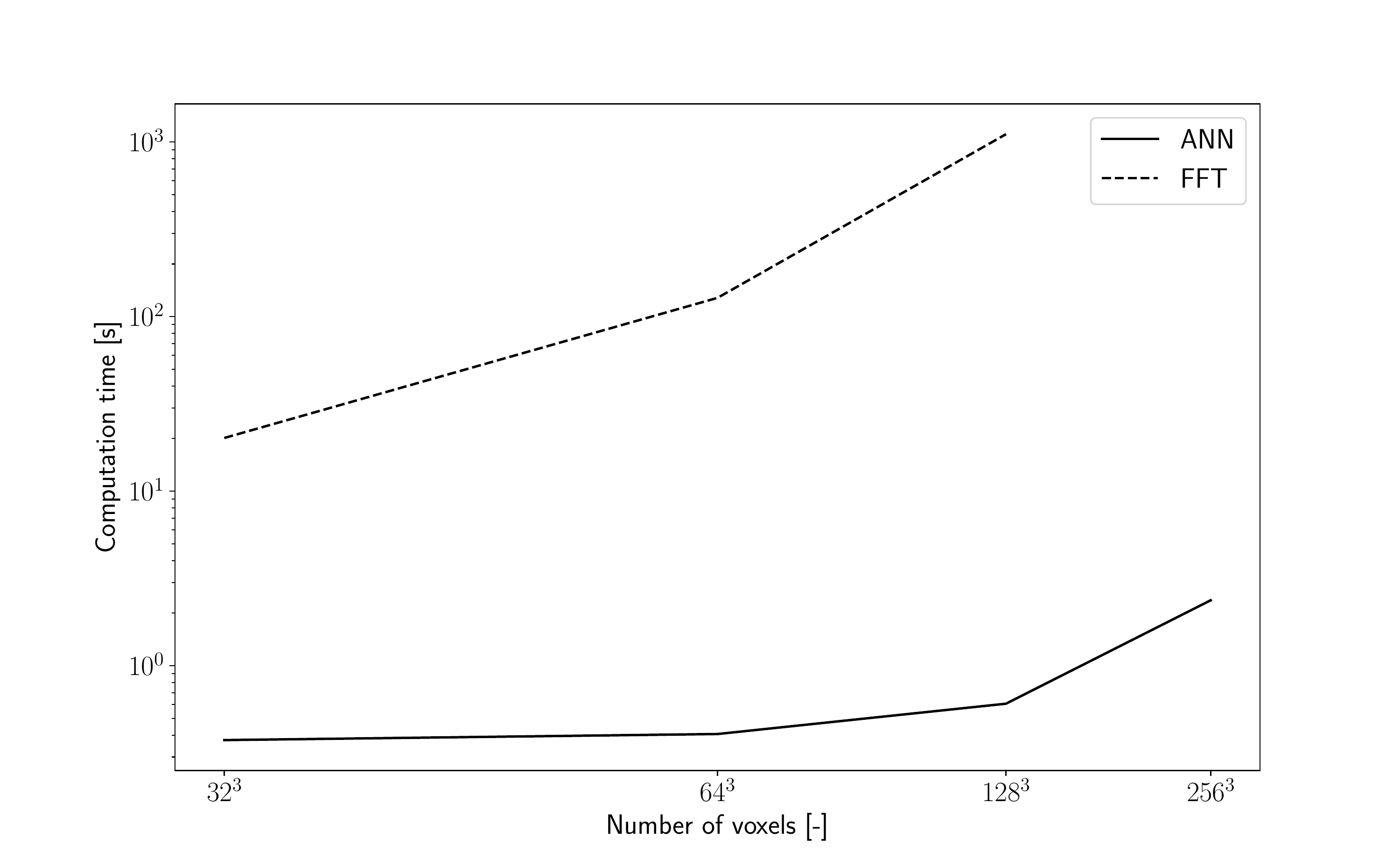}} \caption{Example 2: Comparison of
    computation time for full homogenization in six strain directions of the ANN
model (ANN) with FFT (FFT). For FFT, there are no results for unit cells larger
than $\xi_1 = \xi_2 = \xi_3 = 128^3$ voxels per dimension, as defined in
\autoref{eq:microstructure}, due to memory limitations of the GPU. All
computations were carried out on a mobile workstation with Nvidia M5000M GPU.}
\label{fig:time} \end{figure}

Beside the accuracy comparison, the computation time for a full homogenization
is investigated in \autoref{fig:time}. The evaluation time on a mobile
workstation with Nvidia M5000M GPU for the deterministic FFT used by the MC
reference solution and the ANN is plotted against the number of voxels. Both
methods use GPU acceleration, the FFT algorithm utilizes \textit{CuPy}
\cite{cupy_learningsys2017} and the ANN algorithm \textit{Tensorflow}
\cite{tensorflow2015-whitepaper}. It can be seen, that the ANN approach is
magnitudes faster than FFT. Furthermore, the scaling and memory efficiency of
the ANN with respect to the number of used voxels is better than in the case of
FFT. Using e.g. $\xi_1 = \xi_2 = \xi_3 = 128^3$ voxels from
\autoref{eq:microstructure} yields the memory limit for the FFT approach on the
8 GB GPU, whereas the ANN is capable of processing up to $\xi_1 = \xi_2 = \xi_3
= 256^3$ voxels. With this speed up it is possible to carry out uncertain
computations without a large scale computer cluster system.

 \subsection{Example 3: effective transversely isotropic properties of carbon
fiber reinforced polymer with ANN trained on spherical inclusions}
\label{sec:example_3} \subsubsection{Problem description} 
\begin{figure}[htb] \centering
    \subfloat{\includegraphics[width=\textwidth]{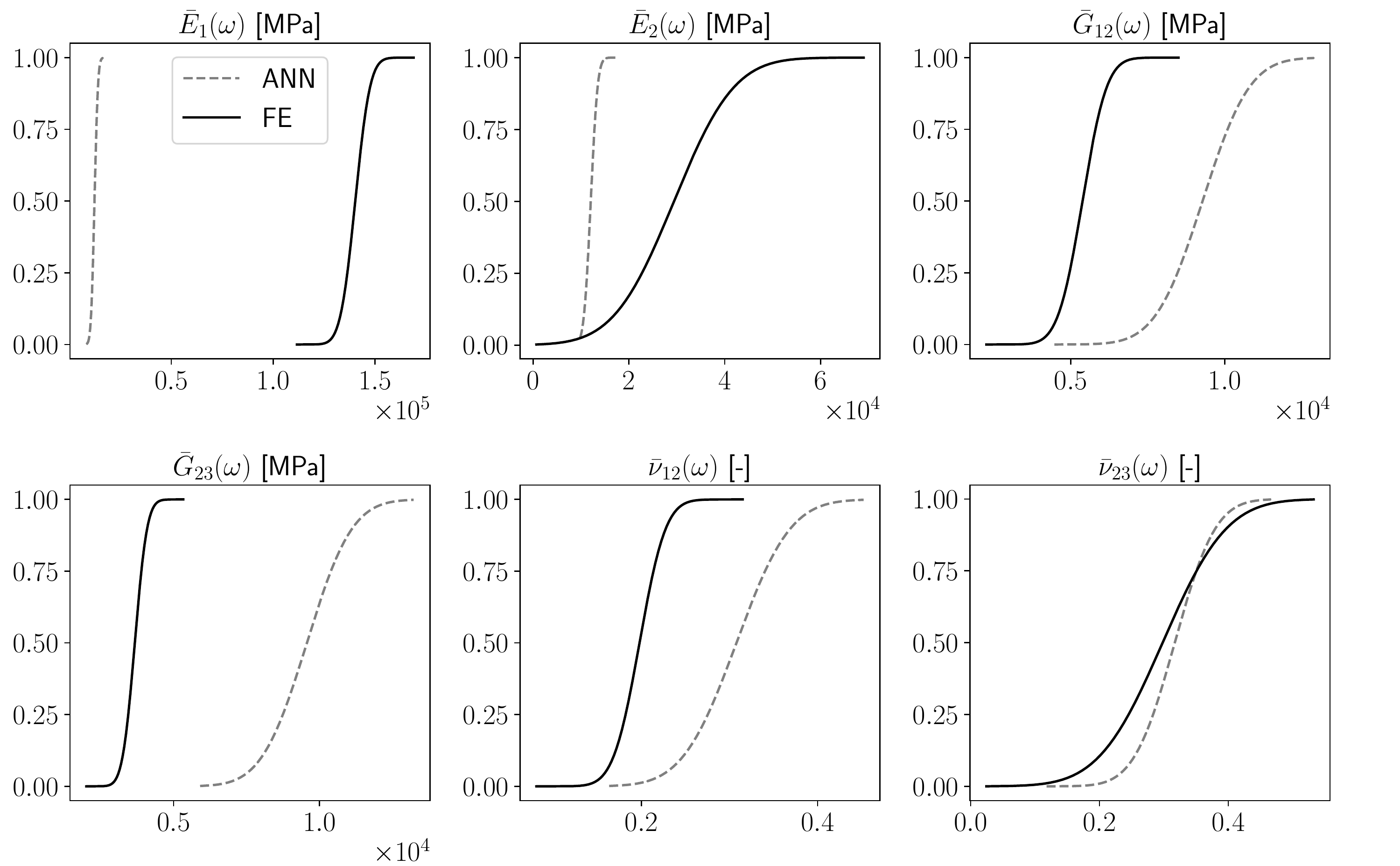}} \caption{Example
    3: Comparison of CDFs of ANN model (ANN) trained on spherical inclusions
from Example 2 and \cite{caylak_polymorphic_nodate} (FE). }
    \label{fig:s_w_s} \end{figure}

In Example 3, the ANN
from Example \ref{sec:example_2}, which is trained on spherical inclusions in
\ah{Figure} \autoref{fig:sphere_rve}, is tested on the microstructure from
Example
\ref{sec:example_1}, which are cylindrical single fiber inclusions, seen in
\ah{Figure} \autoref{fig:single_rve}. The cylindrical single fiber inclusions
inherits
transversely isotropic macroscopic material behaviour, while the spherical
inclusions are isotropic. The aim of this example is to investigate the
generalization behaviour of the ANN with respect to different microstructures.
As can be seen in \autoref{fig:s_w_s}, large deviations of the ANN predictions
to the finite element reference solution are present. The ANN trained on
spherical inclusions with isotropic material behaviour is not able to capture
the transversely isotropic behaviour of the single fiber inclusions. This
indicates, that the ANN is only capable to homogenize microstructures it is
trained on. 

\begin{figure}[htb] \centering
    \subfloat{\includegraphics[width=\textwidth]{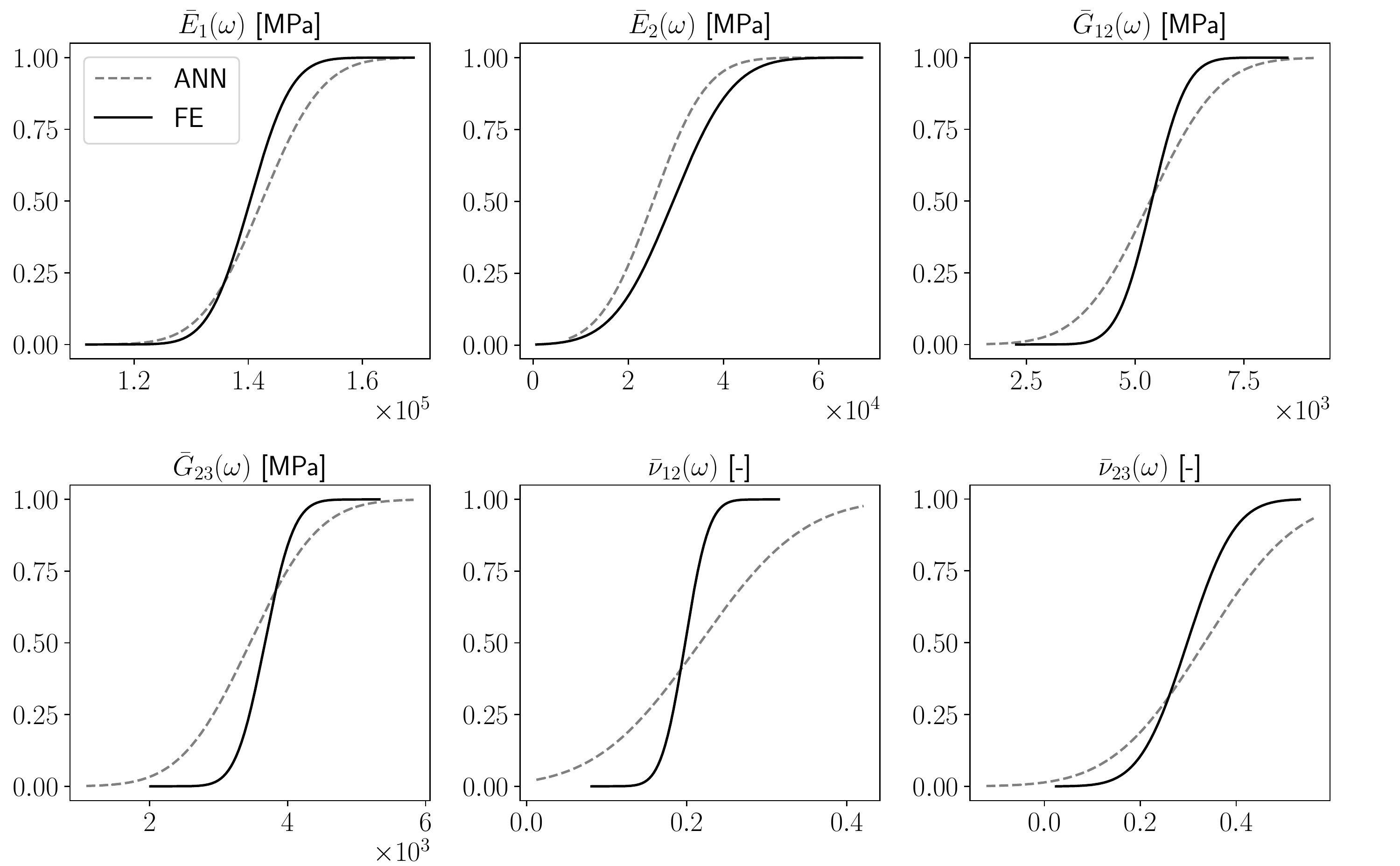}}
    \caption{Example 3: Comparison of CDFs of ANN model (ANN) trained on single
    fiber and spherical inclusions from Example 1 and Example 2, respectively,
and \cite{caylak_polymorphic_nodate} (FE). } \label{fig:cross_single}
\end{figure}

\begin{figure}[htb] \centering \subfloat{\includegraphics[width=\textwidth,
    height=0.5\textwidth]{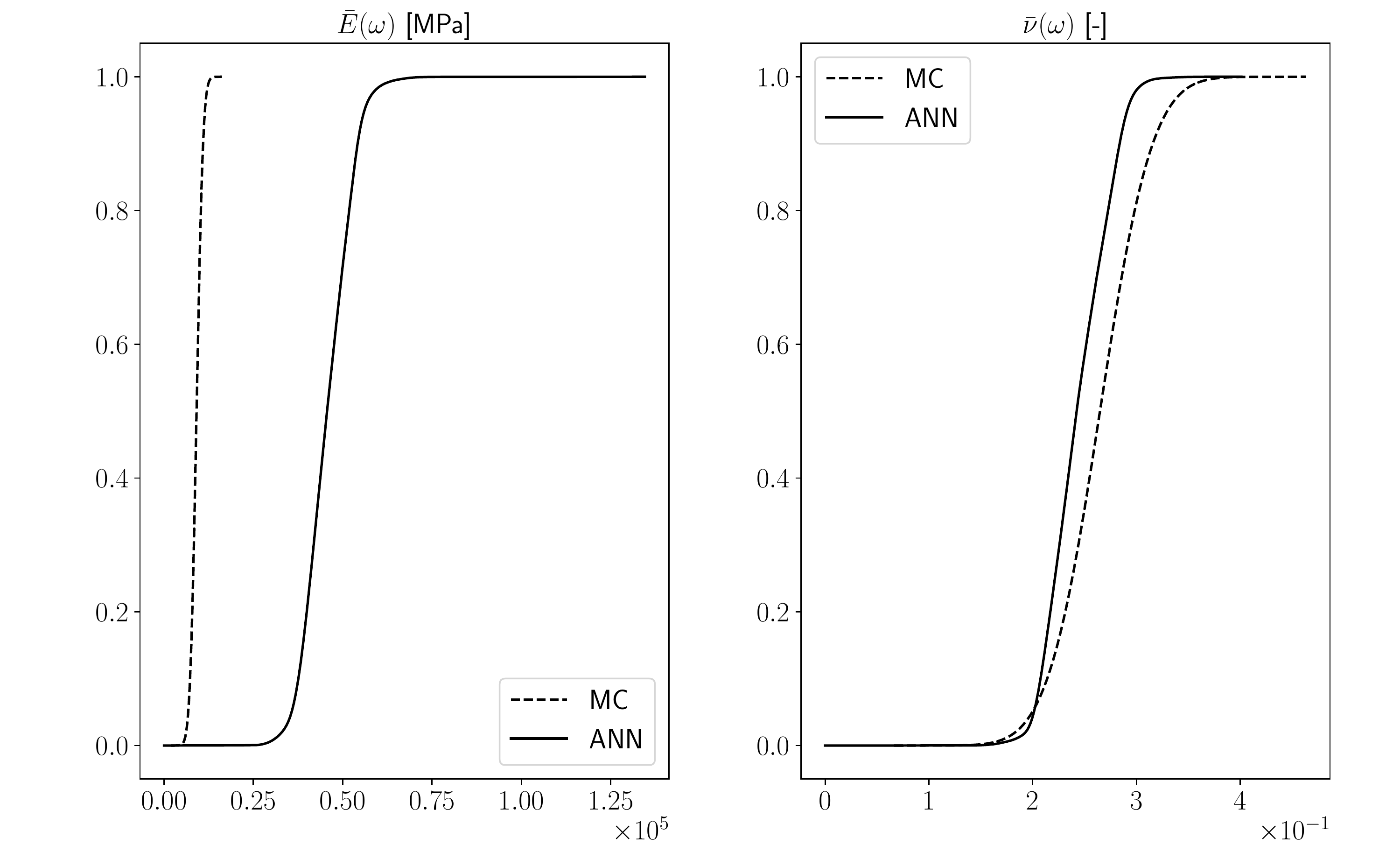}} \caption{Example 3: Comparison of
    CDFs of ANN model (ANN) trained on single fiber and spherical inclusions
from Example 1 and Example 2, respectively, and $10^3$ MC simulations. }
\label{fig:cross_sphere} \end{figure}

To investigate the performance on a mixed set with respect to the underlying
microstructure, the two datasets from Example 1 and Example 2, in the following
denoted by $\mathbb{D}_{ANN}^{(1)}$ and $\mathbb{D}_{ANN}^{(2)}$, respectively,
are simply combined, such that \begin{equation*} \mathbb{D}_{ANN} =
\mathbb{D}_{ANN}^{(1)} + \mathbb{D}_{ANN}^{(2)} \end{equation*} in an
appropriate sense. Then, keeping the hyperparameters and topology of the ANN the
same as in Example 2, namely adopting the hyperparameters in \autoref{tab:hp_2}
and utilizing a \textit{Densenet} for the CNN, the ANN is trained on the mixed
set $\mathbb{D}_{ANN}$. After training, UQ is carried out twice, once with
respect to the configurations in Example 1 and once with respect to the
configurations in Example 2. The resulting CDFs are shown in
\autoref{fig:cross_single} and \autoref{fig:cross_sphere}, respectively. 

First consider the results from \autoref{fig:cross_single}. Here, the
predictions with respect to the single fiber microstructure from Example 1 of
the ANN trained on the mixed set are shown. Compared to \autoref{fig:s_w_s},
where the ANN is trained on spherical inclusions, the results have improved and
show better agreement with the reference solution. The quality of the outcomes,
i.e. the overlap between ANN and reference solution, are even comparable with
the results from Example 1 in \autoref{fig:caylak}, where the ANN is trained
purely on single fibers. The results indicate that the ANN is able to correctly
identify the nature of the microstructure, given that it is included in the
training set.

Second consider the results from \autoref{fig:cross_sphere}. Here the
predictions with respect to the spherical inclusion microstructure from Example
2 of the ANN trained on the mixed set is shown. Deviations of the macroscopic
Young's modulus can be observed and the predictive capability is worse in
comparison with the ANN from Example 2, which was trained purely on spherical
microstructures. 

The authors are aware of this discrepancy, as it is expected for the ANN to
perform better on microstructures, for which more samples are available in the
training dataset. Currently, no solution to this open problem is available.

\section{Conclusion and outlook} \label{sec:Summary_Outlook} The objective of
this paper is to present a deep learning driven pseudospectral PCE based FFT
homogenization algorithm in order to quantify efficiently uncertainties in
effective properties of complex, three dimensional microstructures. Here,
uncertainties arising from material parameters of single constituents as well as
the geometry of the underlying microstructure were considered. In order to
reduce the computational effort of uncertain full-field homogenization, which is
commonly treated by FEM discretization and MC methods, a deep learning algorithm
is proposed. The ANN is trained on FFT homogenized samples to circumvent
meshing, which leads to a deterministic surrogate. Then, by usage of
pseudospectral PCE, a stochastic surrogate is established, capable of predicting
uncertain effective properties in multiple loading directions. Therefore, the
proposed approach is able to calculate the full uncertain effective elasticity
tensor.

Several examples were given. The first example shows the ability of the proposed
algorithm to predict uncertain effective properties of transversely linear
elastic \ah{carbon fiber reinforced polymers} by comparing it to established 
methods from the literature. Here,
the presented results indicate a good agreement with
\cite{caylak_polymorphic_nodate}. The second example expands the method further
to complex microstructures, comparing MC of FFT and with the proposed PCE/ANN 
algorithm.
Again, the proposed algorithm is capable of predicting uncertain effective
properties. \ah{It is shown, that the error induced from the PCE is small in
    comparison with MC, indicating that the main error source is the error by
    the ANN. Additionally, a significant speed up in the deterministic solution 
    by the ANN
compared to FFT was reported.} The third example shows the problem of
generalization of the ANN predictions. Here, an ANN trained sole on one type of
microstructure was not able to transfer the knowledge to other microstructures.
\ah{This means, that if one wants general prediction capabilities, a broad 
    family of
diverse microstructures have to be included in the training set during training 
stage. This could lead to large training sets, as for every
type of microstructure thousands of different samples are needed. While the
proposed approach is well suited for a specific family of microstructures, which
are included in the training stage,
recent developments in operator learning \cite{lu2021learning} could permit 
more generalization power, as reported in e.g. \cite{lin2021operator} and 
\cite{ranade2021generalized}.}

Compared to different approaches in the literature, this algorithm is capable of
predicting effective properties projected to uncertainties of material
parameters and geometry in three dimensional microstructures, were attempts in
the literature were restricted to either deterministic approaches, two
dimensions, fixed material parameters or a single loading direction.

The proposed approach enabled full-field uncertainty quantification by building
an efficient surrogate based on deep learning. Here, the computational heavy
part is front loaded in the sense, that the single model evaluation is cheap in
comparison to FEM or FFT, but the training, or preprocessing, is more involved,
which on the other hand, is only a one time effort. Nevertheless, the reduction
of sample generation and training time is an important topic for future work. An
interesting approach in this context is the utilization of so called
\textit{physics informed neural networks} \cite{raissi_physics-informed_2019},
which solve the underlying system of partial differential equations directly,
effectively bypassing the need of sample creation and label generation. An
intrusive Galerkin projection, as presented in \cite{caylak_polymorphic_nodate},
could be a possibility to include uncertainties in such an approach.
Furthermore, real world examples, such as complex microstructures from CT-scans,
and experimental data are needed to validate the proposed method. 

 \clearpage

\section*{Acknowledgement} The support of the research in this work by the
German ``Ministerium für Kultur und Wissenschaft des Landes NRW'' is gratefully
acknowledged. The authors gratefully acknowledge the funding of this project by
computing time provided by the Paderborn Center for Parallel Computing (PC2).
The financial support of this research by the  "DFG-Schwerpunktprogramm SPP1886"
is gratefully acknowledged.

\bibliography{literature_henkes} \bibliographystyle{elsarticle-num}
\end{document}